\documentclass{article}

\PassOptionsToPackage{numbers, sort&compress}{natbib}
\usepackage[final]{neurips_2019}
\usepackage{tchdr}

\usepackage[utf8]{inputenc} 
\usepackage[T1]{fontenc}    
\usepackage{hyperref}       
\usepackage{url}            
\usepackage{booktabs}       
\usepackage{amsfonts}       
\usepackage{nicefrac}       
\usepackage{microtype}      

\newcommand{\prox}[2]{\ensuremath{\operatorname{prox}_{#1}\left(#2\right)}}

\newdimen{\algindent}
\makeatletter
\newcommand{\LineComment}[2][0]{\Statex \hspace{#1\algindent} \hskip\ALG@thistlm $\triangleright$ #2}
\makeatother

\title{Sparse Variational Inference:\\ Bayesian Coresets from Scratch}

\author{%
  Trevor Campbell\\
  Department of Statistics\\
  University of British Columbia\\
  Vancouver, BC V6T 1Z4 \\
  \texttt{trevor@stat.ubc.ca} \\
\And
  Boyan Beronov\\
  Department of Computer Science\\
  University of British Columbia\\
  Vancouver, BC V6T 1Z4\\
  \texttt{beronov@cs.ubc.ca}
}

\begin{document}

\maketitle
\begin{abstract}
The proliferation of automated inference algorithms in Bayesian statistics
has provided practitioners newfound access to fast, reproducible data analysis and powerful statistical models. 
Designing automated methods that are also both computationally scalable 
and theoretically sound, however, remains a significant challenge. 
Recent work on \emph{Bayesian coresets} takes the approach of compressing the dataset
 before running a standard inference algorithm, providing both scalability and guarantees on posterior approximation error.
But the automation of past coreset methods is limited because they depend on the availability of a reasonable coarse posterior approximation,
which is difficult to specify in practice.
In the present work we remove this requirement by formulating coreset
construction
as sparsity-constrained variational inference within an exponential family.
This perspective leads to a novel construction via greedy optimization, and also provides
a unifying information-geometric view of present and past methods. 
The proposed \emph{Riemannian coreset construction} algorithm is fully automated, 
requiring no problem-specific inputs aside from the probabilistic model and dataset. 
In addition to being significantly easier to use than past methods, experiments demonstrate that past coreset constructions are 
fundamentally limited by the fixed coarse posterior approximation; in contrast, the proposed 
algorithm is able to continually improve the coreset, providing state-of-the-art Bayesian dataset summarization with
orders-of-magnitude reduction in KL divergence to the exact posterior.
\end{abstract}

\section{Introduction}
Bayesian statistical models are  powerful tools for learning from data, with the ability to
encode complex hierarchical dependence and domain expertise, as well as coherently
quantify uncertainty in latent parameters.  In practice, however, 
exact Bayesian inference is typically intractable, and we must use approximate inference algorithms
such as Markov chain Monte Carlo (MCMC) [\citealp{Robert04}; \citealp[Ch.~11,12]{Gelman13}] and variational inference 
(VI) \citep{Jordan99,Wainwright08}. Until recently, implementations of these methods were 
created on a per-model basis, requiring expert input to design the MCMC transition kernels 
or derive VI gradient updates. But developments in automated tools---e.g.,~automatic differentiation \citep{Baydin18,Kucukelbir17},
``black-box'' gradient estimates \citep{Ranganath14}, and Hamiltonian transition kernels \citep{Neal11,Hoffman14}---have obviated much of this expert input, 
greatly expanding the repertoire of Bayesian models accessible to practitioners.


In modern data analysis problems, automation alone is insufficient; inference algorithms must also be computationally scalable---to
handle the ever-growing size of datasets---and provide theoretical guarantees on the quality of their output such that statistical pracitioners may
confidently use them in failure-sensitive settings.
Here the standard set of tools falls short. Designing correct MCMC schemes in the large-scale data setting is a challenging, problem-specific task \citep{Bardenet15,Scott16,Betancourt15};
and despite recent results in asymptotic theory \citep{Alquier18,Wang18,Yang18,CheriefAbdellatif18}, it is difficult to assess the effect of the variational family
on VI approximations for finite data, where a poor choice can result in severe underestimation of posterior uncertainty \citep[Ch.~21]{Murphy12}. 
Other scalable Bayesian inference algorithms have largely been developed by modifying standard inference algorithms to handle distributed
or streaming data processing
\citep{Hoffman13,Scott16,Rabinovich15,Broderick13,Campbell15,Welling11,Ahn12,Bardenet14,Korattikara14,Maclaurin14,Bardenet15,Srivastava15,Entezari16,Angelino16}, 
which tend to have no guarantees on inferential quality and  require extensive model-specific expert tuning.


Bayesian coresets (``core of a dataset'') \citep{Huggins16,Campbell17,Campbell18} are an alternative approach---based on the notion that large datasets often contain 
a significant fraction of redundant data---that summarize and sparsify the data as a preprocessing step before running a standard
inference algorithm such as MCMC or VI. In contrast to other large-scale inference 
techniques, Bayesian coreset construction is computationally inexpensive, simple to implement, 
and provides theoretical guarantees relating coreset size to posterior approximation quality. 
However, state-of-the-art algorithms formulate coreset construction as a sparse regression problem in a Hilbert space, which
involves the choice of a weighted $L^2$ inner product \citep{Campbell17}.
If left to the user, the choice of weighting distribution significantly reduces the overall automation of the approach; and current methods for
finding the weighting distribution programatically are generally as expensive as posterior inference on the full dataset itself. 
Further, even if an appropriate inner product is specified, computing it exactly is typically intractable, requiring the use of finite-dimensional
projections for approximation \citep{Campbell17}. Although the problem in finite-dimensions can be studied using well-known techniques
from sparse regression, compressed sensing, random sketching, boosting, and greedy approximation
\citep{Clarkson10,LacosteJulien15,Locatello17,Barron08,Chen10,Schapire90,Huszar12,Freund97,Candes05,Candes07,Donoho06,Boche15,Mallat93,Chen89,Chen99,Tropp04,Tibshirani96,Geppert17,Ahfock17}, these projections incur an unknown error in the construction process in practice,
 and preclude asymptotic consistency as the coreset size grows.


In this work, we provide a new formulation of coreset construction as exponential family variational inference with a sparsity constraint.
The fact that coresets form a sparse subset of an exponential family is crucial in two regards. First, it enables tractable unbiased 
Kullback-Leibler (KL) divergence gradient estimation, which is used in the development of a novel coreset construction algorithm based on greedy optimization.
In contrast to past work, this algorithm is fully
automated, with no problem-specific inputs aside from the probabilistic model and dataset. 
Second, it provides a unifying view and strong theoretical underpinnings of both the present and past coreset constructions through Riemannian information geometry.
In particular, past methods are shown to operate in a single tangent space of
the coreset manifold; our experiments show that this fundamentally limits the quality of the coreset constructed with these methods.
In contrast, the proposed method proceeds along the manifold towards the posterior target, and is able to continually improve its approximation. 
Furthermore,
new relationships between the optimization objective of past approaches and the coreset posterior KL divergence are derived.
The paper concludes with experiments demonstrating that, compared with past methods, \emph{Riemannian coreset construction} is both easier to use 
and provides orders-of-magnitude reduction in KL divergence to the exact posterior.

\section{Background}
In the problem setting of the present paper, we are given a probability density $\pi(\theta)$ for variables $\theta\in\Theta$
that decomposes into $N$ potentials $(f_n(\theta))_{n=1}^N$ and a base density $\pi_0(\theta)$,
\[
\pi(\theta) &\defined \frac{1}{Z}\exp\left(\sum_{n=1}^N f_n(\theta)\right)\pi_0(\theta),
\]
where $Z$ is the (unknown) normalization constant.
Such distributions arise frequently in a number of scenarios: for example, in Bayesian statistical inference problems
with conditionally independent data given $\theta$, the functions $f_n$ are the log-likelihood terms
for the $N$ data points, $\pi_0$ is the prior density, and $\pi$ is the posterior; or in undirected graphical models, the functions $f_n$ and $\log\pi_0$ might represent $N+1$ potentials.
The algorithms and analysis in the present work are agnostic to their particular meaning, but for clarity we 
will focus on the setting of Bayesian inference throughout.

As it is often intractable to compute expectations under $\pi$ exactly, 
practitioners have turned to approximate algorithms. Markov chain Monte Carlo (MCMC) methods \citep{Robert04,Neal11,Hoffman14}, which return approximate samples from $\pi$, 
remain the gold standard for this purpose. But since each sample typically requires at least one evaluation of a function proportional to $\pi$ with computational cost $\Theta(N)$,
in the large $N$ setting it is expensive to obtain sufficiently many samples to provide high confidence in empirical estimates. 
To reduce the cost of MCMC, we can instead run it on
a small, weighted subset of data known as a \emph{Bayesian coreset}
\citep{Huggins16}, a concept originating from the computational geometry and
optimization literature
\citep{Agarwal05,Langberg10,Feldman11,Feldman13,Bachem17,Braverman16}.
Let $w \in \reals_{\ge 0}^N$ be a sparse vector of nonnegative weights such that
only $M\ll N$ are nonzero, i.e.~$\|w\|_0 \defined \sum_{n=1}^N\ind\left[w_n>0\right] \leq M$.
Then we approximate the full log-density with a $w$-reweighted sum with normalization $Z(w) > 0$ 
and run MCMC on the approximation\footnote{Throughout, $[N] \defined \{1,\dots, N\}$, $1$ and $0$ are the constant vectors of all 1s / 0s respectively (the dimension will be clear from context), $1_\mcA$ is the indicator vector for $\mcA\subseteq [N]$, and $1_n$ is the indicator vector for $n\in[N]$.},
\[
\pi_w(\theta) &\defined \frac{1}{Z(w)} \exp\left(\sum_{n=1}^Nw_n
f_n(\theta)\right)\pi_0(\theta),
\]
where $\pi_1 = \pi$ corresponds to the full density.
If $M \ll N$, evaluating a function proportional to $\pi_w$ is much less
expensive than doing so for the original $\pi$, resulting in a significant reduction in MCMC computation time. 
The major challenge posed by this approach, then, is to find a set of weights $w$ 
that renders $\pi_w$ as close as possible to $\pi$ while maintaining sparsity.
Past work \citep{Campbell17,Campbell18} formulated this as a sparse regression  problem in a Hilbert space
with the $L^2(\hpi)$ norm for some weighting distribution $\hpi$
and vectors\footnote{In \citep{Campbell17}, the $\EE_{\hpi}\left[f_n\right]$ term was missing; it is necessary to account for the shift-invariance of potentials.} $g_n \defined \left(f_n - \EE_{\hpi}\left[f_n\right]\right)$,
\[
w^\star = \argmin_{w\in\reals^N} \,\, \EE_{\hpi}\left[ \left( \sum_{n=1}^Ng_n - \sum_{n=1}^Nw_ng_n\right)^2\right] \,\, \text{s.t.}\,\, w\geq 0,\,\, \|w\|_0\leq M.\label{eq:hilbertcoresets}
\]
As the expectation is generally intractable to compute exactly, a Monte Carlo approximation is used in its place:
taking samples $(\theta_s)_{s=1}^S \distiid \hpi$ and setting 
$\hg_n = \sqrt{S}^{-1}\left[g_n(\theta_1)-\bg_n, \dots, g_n(\theta_S)-\bg_n\right]^T\in\reals^S$ 
where $\bg_n = \frac{1}{S}\sum_{s=1}^S g_n(\theta_s)$
yields a 
linear finite-dimensional sparse regression problem in $\reals^S$,
\[
w^\star &= \argmin_{w\in\reals^N} \,\,\left\|\sum_{n=1}^N\hg_n - \sum_{n=1}^Nw_n\hg_n\right\|^2_2 \,\, \text{s.t.}\,\, w\geq 0,\,\, \|w\|_0\leq M,\label{eq:approxhilbertcoresets}
\]
which can be solved with sparse optimization techniques \citep{Mallat93,Chen89,Tropp04,Barron08,Campbell17,Campbell18,Frank56,Guelat86,Jaggi13,LacosteJulien15,Locatello17}.
%
%
However, there are two drawbacks inherent to the Hilbert space formulation. First, the
use of the $L^2(\hpi)$ norm requires the selection of the weighting function $\hpi$, posing a barrier to 
the full automation of coreset construction. There is currently no guidance on how to select $\hpi$,
or the effect of different choices in the literature. We show in \cref{sec:infogeometry,sec:expts}
that using such a fixed weighting $\hpi$ fundamentally limits the quality of coreset construction.
Second, the inner products typically cannot be
computed exactly, requiring a Monte Carlo approximation.  This adds noise to the construction and precludes 
asymptotic consistency (in the sense that $\pi_w \not\to \pi_1$ as the sparsity budget $M\to \infty$).
Addressing these drawbacks is the focus of the present work.

\section{Bayesian coresets from scratch} \label{sec:sparsevi}
In this section, we provide a new formulation of Bayesian coreset construction as variational
inference over an exponential family with sparse natural parameters,
and develop an iterative greedy algorithm for optimization.

\subsection{Sparse exponential family variational inference}
We formulate coreset construction as a sparse variational inference problem,
\[
w^\star = \argmin_{w\in\reals^N} \quad & \kl{\pi_w}{\pi_1} \quad \text{s.t.} \quad w\geq 0,\,\, \|w\|_0 \leq M.
\label{eq:sparsekl}
\]
Expanding the objective and denoting expectations under $\pi_w$ as $\EE_w$,
\[
 \kl{\pi_w}{\pi_1} = \log Z(1) - \log Z(w) - \sum_{n=1}^N (1-w_n)\EE_w\left[f_n(\theta)\right].\label{eq:kldiv}
\]
\cref{eq:kldiv} illuminates the major challenges with the variational approach posed in \cref{eq:sparsekl}. First, the normalization constant $Z(w)$ of $\pi_w$---itself 
a function of the weights $w$---is unknown; typically, the form of the approximate 
distribution is known fully in variational inference. Second, even if the constant were known, computing the objective in \cref{eq:sparsekl}
requires taking expectations under $\pi_w$, which is in general just as difficult as the original
problem of sampling from the true posterior $\pi_1$. 

Two key insights in this work address these issues and lead to both the development of a new coreset construction 
algorithm (\cref{alg:greedy}) and a more comprehensive understanding of 
the coreset construction literature (\cref{sec:infogeometry}).
First, the coresets form a sparse subset of an exponential family: 
the nonnegative weights form the natural parameter $w\in\reals_{\ge 0}^N$,
the component potentials $(f_n(\theta))_{n=1}^N$ form the sufficient statistic,
$\log Z(w)$ is the log partition function,
and $\pi_0$ is the base density,
\[
\pi_w(\theta) &\defined \exp\left(w^T f(\theta) - \log Z(w)\right)\pi_0(\theta) & f(\theta) &\defined \left[\begin{array}{ccc} f_1(\theta) & \dots & f_N(\theta)\end{array}\right]^T.\label{eq:exponentialfamily}
\]
Using the well-known fact that the gradient of an exponential family
log-partition function is the mean of the sufficient statistic,
$\EE_w\left[f(\theta)\right] = \grad_w\log Z(w)$,
we can rewrite the optimization \cref{eq:sparsekl} as
\[
w^\star = \argmin_{w\in\reals^N} \quad  \log Z(1) - \log Z(w) -
(1-w)^T\grad_w\log Z(w) \quad \text{s.t.} \quad  w\geq 0,\,\,\|w\|_0 \leq M.
\label{eq:sparsekl2}
\]
Taking the gradient of this objective function
and noting again that, for an exponential family, the Hessian of the log-partition function $\log Z(w)$  is the covariance of the sufficient statistic,
\[
\grad_w \kl{\pi_w}{\pi_1} &= -\grad_w^2\log Z(w) (1-w) = -\cov_w\left[f,
f^T(1-w)\right],
\label{eq:klgrad}
\]
where $\cov_w$ denotes covariance under $\pi_w$. In other words, increasing the weight $w_n$ by a small amount
decreases $\kl{\pi_w}{\pi_1}$ by an amount proportional to the covariance of the $n^\text{th}$ potential $f_n(\theta)$ with the residual error $\sum_{n=1}^N f_n(\theta) - \sum_{n=1}^Nw_n f_n(\theta)$
under $\pi_w$. If required, it is not difficult to use the connection between derivatives of $\log Z(w)$
and moments of the sufficient statistic under $\pi_w$ to derive $2^\text{nd}$ and higher order derivatives of $\kl{\pi_w}{\pi_1}$.

This provides a natural tool for optimizing the coreset construction objective in \cref{eq:sparsekl}---Monte 
Carlo estimates of sufficient statistic moments---and enables
coreset construction without both the problematic selection of a Hilbert space (i.e., $\hpi$) and finite-dimensional projection error from past approaches.  
But obtaining Monte Carlo estimates requires sampling from $\pi_w$; the second key insight in this work is that as long as we build up the sparse approximation $w$ incrementally,
the iterates will themselves be sparse. Therefore, using a standard Markov chain Monte Carlo algorithm \citep{Hoffman14}
to obtain samples from $\pi_w$ for gradient estimation is actually not expensive---with cost $O(M)$ instead of $O(N)$---despite the potentially
complicated form of $\pi_w$. 

\subsection{Greedy selection}\label{sec:greedyselection}
One option to build up a coreset incrementally is to use a greedy approach (\cref{alg:greedy}) to select and subsequently reweight a single potential function at a time. 
For greedy selection, the na\"ive approach is to select the potential that provides the largest local decrease in KL divergence around the current weights $w$,
i.e., selecting the potential with the largest covariance with the residual error per \cref{eq:klgrad}.
However, since the weight $w_{n^\star}$ will then be optimized over $[0, \infty)$, the selection of the next potential to add
should be invariant to scaling each potential $f_{n}$ by any positive constant. Thus we propose the use of 
the correlation---rather than the covariance---between 
$f_n$ and the residual error $f^T(1-w)$ as the selection criterion:
\[
\hspace{-.2cm}n^\star\!\! = \argmax_{n\in[N]}
\left\{\begin{array}{ll} 
\left|\corr_w\left[f_n, f^T(1-w)\right]\right| &\!\! w_n > 0\\
\corr_w\left[f_n, f^T(1-w)\right]  &\!\! w_n = 0
\end{array}\right..\label{eq:greedyselection}
\]
Although seemingly ad-hoc, this modification will be placed on a solid information-geometric theoretical 
foundation in \cref{prop:riemann} (see also \cref{eq:corrjust} in \cref{sec:riemannproofs}).
Note that since we do not have access to the exact correlations, we must use Monte Carlo estimates via sampling from $\pi_w$ for greedy selection.
Given $S$ samples $(\theta_s)_{s=1}^S\distiid \pi_w$, these are given by the $N$-dimensional vector
\[
 \widehat{\corr}&=
 \diag\left[\frac{1}{S}\sum_{s=1}^S\hg_s\hg_s^T\right]^{-\frac{1}{2}}
 \!\!\!\left(\!\frac{1}{S}\sum_{s=1}^S\hg_s\hg_s^T(1-w)\!\right) &
\hg_s &\defined {\scriptsize \left[\!\!\!\begin{array}{c} f_1(\theta_s) \\
\vdots \\ f_N(\theta_s)\end{array}\!\!\!\right] -
\frac{1}{S}\sum_{r=1}^S\left[\!\!\!\begin{array}{c} f_1(\theta_r) \\ \vdots \\
f_N(\theta_r)\end{array}\!\!\!\right]},\label{eq:fhat}
\]
where $\diag\left[\cdot\right]$ returns a diagonal matrix with the same diagonal entries as its argument.
The details of using the correlation estimate (\cref{eq:fhat}) in the greedy selection rule (\cref{eq:greedyselection}) to add points
to the coreset are shown in lines 4--9 of \cref{alg:greedy}.
Note that this computation has cost $O(NS)$. If $N$ is large enough that computing the entire vectors $\hg_s\in\reals^N$ is cost-prohibitive, 
one may instead compute $\hg_s$ in \cref{eq:fhat} only for indices in $\mcI \cup \mcU$---where $\mcI = \{n \in[N] : w_n >0\}$
is the set of active indices, and $\mcU$ is 
a uniformly selected subsample of $U\in[N]$ indices---and perform greedy selection only within these indices.
%

\subsection{Weight update}\label{sec:weightupdate}
After selecting a new potential function $n^\star$, we add it to the active set of indices $\mcI \subseteq [N]$
and update the weights by optimizing
\[
w^\star = \argmin_{v \in \reals^N} \kl{\pi_v}{\pi} \quad \text{s.t.} \quad
v\geq 0, \quad (1 - 1_{\mcI})^Tv = 0.\label{eq:fullycorrective}
\]
In particular, we run $T$ steps of generating $S$ samples $(\theta_s)_{s=1}^S\distiid \pi_w$, 
computing a Monte Carlo estimate $D$ of the gradient $\grad_w\kl{\pi_w}{\pi_1}$ based on \cref{eq:klgrad},
\[
D \defined -\frac{1}{S}\sum_{s=1}^S \hg_s\hg_s^T(1-w) \in \reals^N \quad \hg_s\text{ as in \cref{eq:fhat}},
\]
and taking a stochastic gradient step $w_n \gets w_n - \gamma_t D_n$ at step $t\in[T]$ for each $n\in\mcI$, 
using a typical learning rate  $\gamma_t \propto t^{-1}$. The details of the weight update step 
are shown in lines 10--15 of \cref{alg:greedy}. As in the greedy selection step, the cost
of each gradient step is $O(NS)$, due to the $\hg_s^T1$ term in the gradient. If $N$ is large
enough that this computation is cost-prohibitive, one can use $\hg_s$ computed only for indices in $\mcI\cup\mcU$,
where $\mcU$ is a uniformly selected subsample of $U\in[N]$ indices.

\begin{algorithm}[t!]
\caption{Greedy sparse stochastic variational inference}\label{alg:greedy}
\begin{algorithmic}[1]
\Procedure{SparseVI}{$f$, $\pi_0$, $S$, $T$, $(\gamma_t)_{t=1}^\infty$,
$M$}
\State $w \gets 0\in\reals^N$, $\mcI \gets \emptyset$
\For{$m=1, \dots, M$}
\LineComment[1]{Take $S$ samples from the current coreset posterior approximation $\pi_w$}
\State $(\theta_s)_{s=1}^S \distiid \pi_w \propto \exp(w^Tf(\theta))\pi_0(\theta)$
\LineComment{Compute the $N$-dimensional potential vector for each sample}
\State $\hf_s \gets f(\theta_s)\in\reals^N$ for $s\in [S]$, and $\bbf \gets \frac{1}{S}\sum_{s=1}^S \hf_s$
\State $\hg_s \gets \hf_s - \bbf$ for $s\in[S]$
\LineComment{Estimate correlations between the potentials and the residual error}
\State $\widehat{\corr} \gets
\diag\left[\frac{1}{S}\sum_{s=1}^S\hg_s\hg_s^T\right]^{-\frac{1}{2}}
\!\!\!\left(\!\frac{1}{S}\sum_{s=1}^S\hg_s\hg_s^T(1-w)\!\right)
\in\reals^{N}$
\LineComment{Add the best next potential to the coreset}
\State $n^\star \gets \argmax_{n\in[N]} |\widehat{\corr}_n|\ind\left[n\in\mcI\right] + \widehat{\corr}_n\ind\left[n\notin\mcI\right]$
\State $\mcI \gets \mcI \cup \{n^\star\}$
\LineComment{Update all the active weights in $\mcI$ via stochastic gradient descent on $\kl{\pi_w}{\pi}$}
\For{$t=1, \dots, T$}
\LineComment{Use samples from $\pi_w$ to estimate the gradient}
\State $(\theta_s)_{s=1}^S \distiid \pi_w\propto \exp(w^Tf(\theta))\pi_0(\theta)$
\State $\hf_s \gets f(\theta_s)\in\reals^N$ for $s\in [S]$, and $\bbf \gets \frac{1}{S}\sum_{s=1}^S \hf_s$
\State $\hg_s \gets \hf_s - \bbf$ for $s\in[S]$
\State $D \gets -\frac{1}{S}\sum_{s=1}^S \hg_s\hg_s^T(1-w)$
\LineComment{Take a stochastic gradient step for active indices in $\mcI$}
\State $w\gets w - \gamma_t I_{\mcI} D$ where $I_{\mcI} \defined \sum_{n\in\mcI} 1_n1_n^T$ is the diagonal indicator matrix for $\mcI$
\EndFor
\EndFor
\State \Return $w$
\EndProcedure
\end{algorithmic}
\end{algorithm}



\section{The information geometry of coreset construction} \label{sec:infogeometry}
\begin{figure}[t!]
\centering\includegraphics[width=0.75\textwidth]{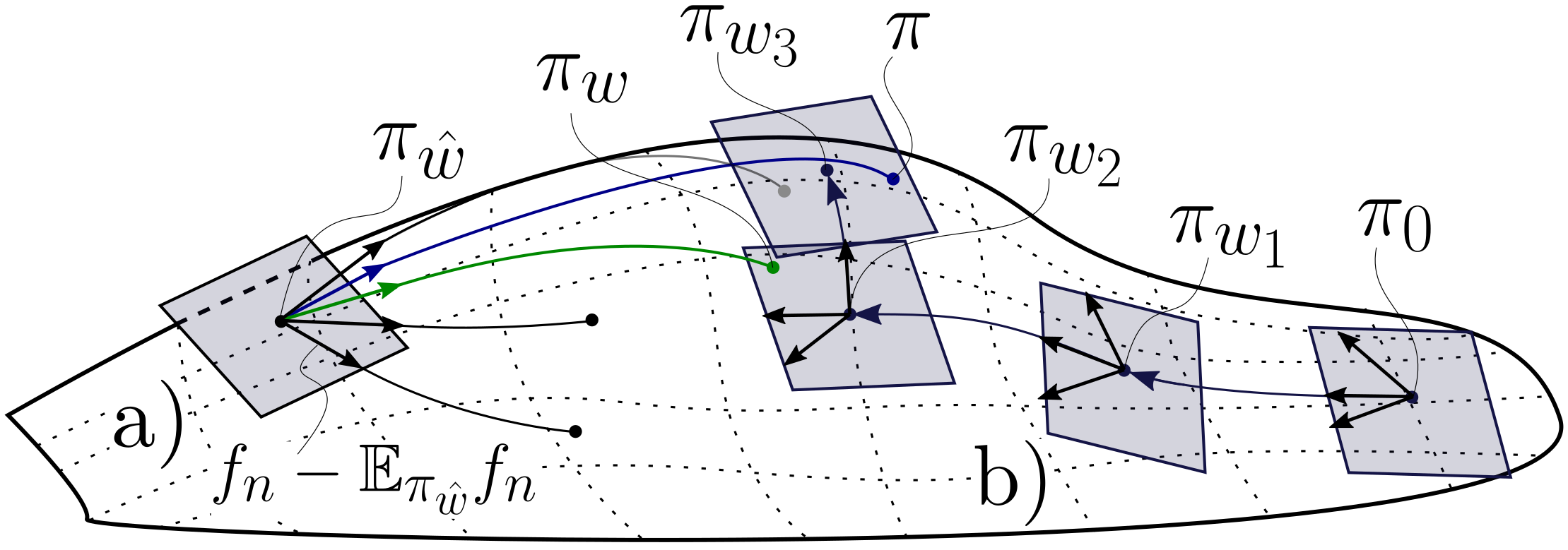}
\caption{Information-geometric view of greedy coreset construction on the coreset manifold $\mcM$. (1a): Hilbert coreset construction, with weighting distribution $\pi_{\hw}$, full posterior $\pi$, coreset posterior $\pi_w$, and arrows denoting initial geodesic directions from $\hw$ towards new datapoints. 
(1b): Riemannian coreset construction, with the path of posterior approximations $\pi_{w_t}$, $t=0,\dots, 3$, and arrows denoting initial geodesic directions towards new datapoints to add
within each tangent plane. }\label{fig:riemann}
\end{figure}

The perspective of coresets as a sparse exponential family also enables the use of information geometry to derive
a unifying connection between the variational formulation and previous constructions. 
In particular, the family of coreset posteriors
defines a Riemannian statistical manifold $\mcM=\{\pi_w\}_{w\in\reals_{\ge
0}^N}$ with chart $\mcM \to \reals_{\ge 0}^N$, endowed with the
\emph{Fisher information metric} $G$ \citep[p.~33,34]{Amari16},
\[
G(w) &= \int \pi_w(\theta) \grad_w\log \pi_w(\theta) \grad_w\log \pi_w(\theta)^T
\dee \theta =\grad_w^2\log Z(w) =  \cov_w\left[f\right].\label{eq:riemannmetric}
\]
For any differentiable curve $\gamma : [0,1]\to\reals_{\ge 0}^N$, the metric
defines a notion of path length,
\[
L(\gamma) = \int_0^1 \sqrt{\der{\gamma(t)}{t}^T G(\gamma(t))\der{\gamma(t)}{t}}\dee t,
\]
and a constant-speed curve of minimal length between any two points
$w,w'\in\reals_{\ge 0}^N$ is referred to as a \emph{geodesic}
\citep[Thm.~5.2]{Amari16}.
The geodesics are the generalization of straight lines in Euclidean space to curved Riemannian manifolds, such as $\mcM$.
Using this information-geometric view, \cref{prop:riemann} shows that
both Hilbert coreset construction (\cref{eq:hilbertcoresets}) and
the proposed greedy sparse variational inference procedure (\cref{alg:greedy})
attempt to directionally align the $\hw \to w$ and $\hw \to 1$ geodesics
on $\mcM$
for $\hw, w, 1\in\reals_{\geq 0}^N$ (reference, coreset, and true posterior weights, respectively)
as illustrated in \cref{fig:riemann}. 
The key difference is that Hilbert coreset construction uses a fixed reference point $\hw$---corresponding to $\hpi$ in 
\cref{eq:hilbertcoresets}---and thus operates entirely in a single tangent space of $\mcM$, while the proposed greedy method uses $\hw = w$
and thus improves its tangent space approximation as the algorithm iterates. For this reason, we refer to the method in \cref{sec:sparsevi}
as a \emph{Riemannian coreset construction} algorithm. In addition to this unification of coreset construction methods, the geometric perspective 
also provides the means to show that the Hilbert coresets objective bounds the 
symmetrized coreset KL divergence $\kl{\pi_w}{\pi}+\kl{\pi}{\pi_w}$ if the Riemannian metric does not vary too much, as shown in \cref{prop:hilbertcoreseterrorguarantees}.
Incidentally, \cref{lem:klest} in \cref{sec:riemannproofs}---which is used to prove \cref{prop:hilbertcoreseterrorguarantees}---also provides a 
nonnegative unbiased estimate of the symmetrized coreset KL divergence, which may be used for performance monitoring in practice.
\bnprop\label{prop:riemann}
Suppose $\hpi$ in \cref{eq:hilbertcoresets} satisfies $\hpi = \pi_{\hw}$ for a
set of weights $\hw\in\reals_{\ge 0}^N$.
For $u,v\in\reals_{\ge 0}^N$, let $\xi_{u\to v}$ denote the initial tangent of
the $u\to v$ geodesic on $\mcM$,
and $\left<\cdot, \cdot\right>_u$ denote the inner product under the Riemannian metric $G(u)$ with induced norm $\|\cdot\|_u$.
Then Hilbert coreset construction in \cref{eq:hilbertcoresets} is equivalent to
\[
w^\star &= \argmin_{w\in\reals^N} \,\,\left\| \xi_{\hw\to 1} - \xi_{\hw \to w}\right\|_{\hw} \quad\text{s.t.}\quad w\geq 0,\,\, \|w\|_0\leq M,
\]
and
each greedy selection step of Riemannian coreset construction in \cref{eq:greedyselection} is equivalent to
\[
n^\star &=\argmin_{n\in[N], t_n\in\reals} \left\| \xi_{w\to 1} -  \xi_{w \to w+t_n 1_n}\right\|_{w}\quad\text{s.t.}\quad \forall\,n\notin\mcI,\,t_n > 0.
\]
\enprop

\bnprop\label{prop:hilbertcoreseterrorguarantees}
Suppose $\hpi$ in \cref{eq:hilbertcoresets} satisfies $\hpi = \pi_{\hw}$ for a
set of weights $\hw\in\reals_{\ge 0}^N$.
Then if $J_{\hpi}(w)$ is the objective function in \cref{eq:hilbertcoresets},
\[
\kl{\pi}{\pi_w} + \kl{\pi_w}{\pi} &\leq C_{\hpi}(w) \cdot J_{\hpi}(w),
\]
where $C_{\hpi}(w) \defined \EE_{U\dist\distUnif[0,1]}\!\left[
\lambda_{\max}\!\left(G(\hw)^{-1/2}G((1-U)w + U1)G(\hw)^{-1/2}\right)
\right]$.
In particular, if $\grad_w^2\log Z(w)$ is constant in $w\in\reals_{\ge 0}^N$,
then $C_{\hpi}(w) = 1$.
\enprop

\section{Experiments}\label{sec:expts}
\renewcommand{\UrlFont}{\ttfamily}

In this section, we compare the quality of coresets constructed via the proposed \texttt{SparseVI} greedy coreset construction 
method, uniform random subsampling, and Hilbert coreset construction (GIGA \citep{Campbell18}). In particular,
for GIGA we used a 100-dimensional random projection generated from a Gaussian $\hpi$ with two parametrizations:
one with mean and covariance set using the moments of the exact posterior (Optimal) which is a benchmark but is not possible to achieve in practice; 
and one with mean and covariance uniformly distributed
between the prior and the posterior with $75\%$ relative noise added (Realistic)
to simulate the choice of $\hpi$ without exact posterior information.
Experiments were performed on a machine with an Intel i7 8700K processor and 32GB memory; code is available at \url{www.github.com/trevorcampbell/bayesian-coresets}.

\subsection{Synthetic Gaussian posterior inference}\label{sec:synthexpt}

\begin{figure}[t!]
\begin{center}
\begin{subfigure}{0.4\textwidth}
\centering\includegraphics[width=\textwidth]{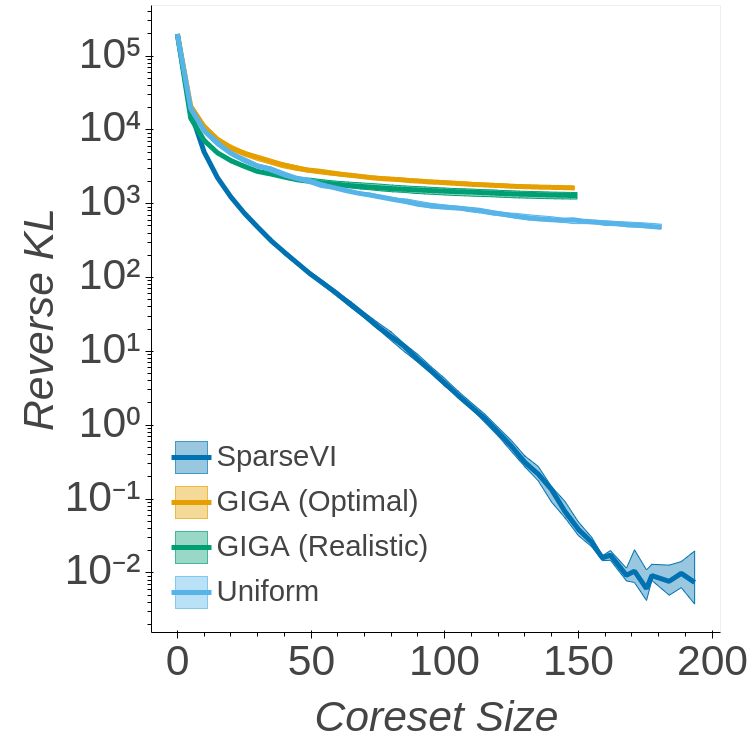}
 \caption{}\label{fig:gaussian_kl}
\end{subfigure}
\begin{subfigure}{0.58\textwidth}
\centering\includegraphics[width=0.95\textwidth]{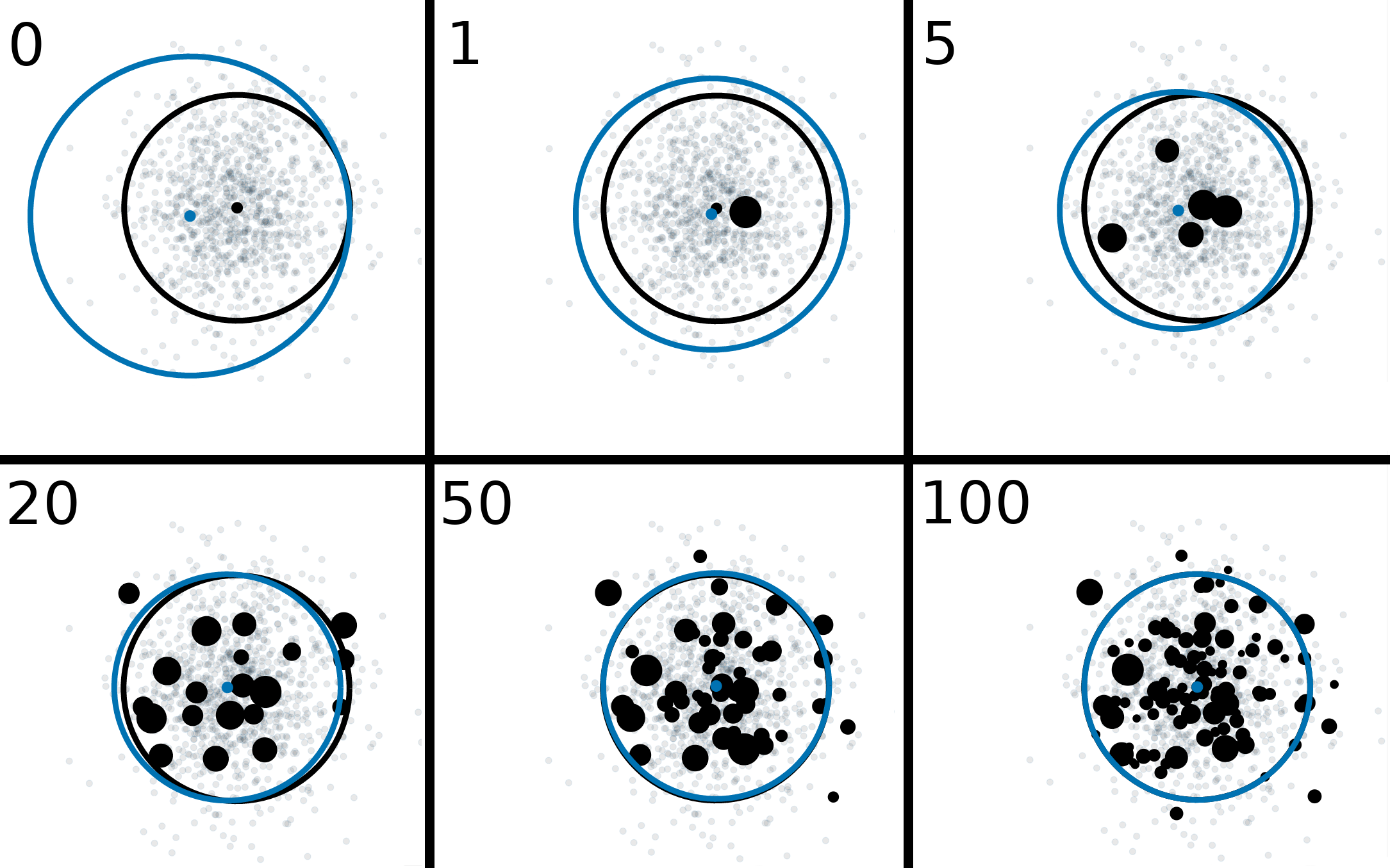}
 \caption{}\label{fig:gaussian_coreset}
\end{subfigure}
\end{center}
\caption{(\ref{fig:gaussian_kl}): Synthetic comparison of coreset construction methods.
Solid lines show the median KL divergence over 10 trials, with $25^\text{th}$ and $75^\text{th}$ percentiles shown by shaded areas.
(\ref{fig:gaussian_coreset}): 2D projection of coresets after 0, 1, 5, 20, 50, and 100 iterations via \texttt{SparseVI}.
True/coreset posterior and $2\sigma$-predictive ellipses are shown in black/blue respectively. Coreset points are black with radius denoting weight.}\label{fig:gaussian}
\vspace{-.2cm}
\end{figure}

We first compared the coreset construction algorithms on a synthetic example involving posterior inference for the mean of a $d$-dimensional Gaussian with Gaussian observations,
\[
\theta &\dist \distNorm(\mu_0, \Sigma_0) & x_n & \distiid \distNorm(\theta, \Sigma), \quad n=1, \dots, N.
\]
We selected this example because it decouples the evaluation of the coreset construction methods from the concerns of stochastic optimization and
approximate posterior inference: the coreset posterior $\pi_w$ is a Gaussian $\pi_w = \distNorm(\mu_w, \Sigma_w)$ with closed-form expressions for the parameters
as well as covariance (see \cref{sec:gaussiancovariance} for the derivation),
\[
\Sigma_w = \big(\Sigma^{-1}_0 + {\textstyle
\sum_{n=1}^N}w_n\Sigma^{-1}\big)^{-1} &\qquad  \mu_w =
\Sigma_w\big(\Sigma^{-1}_0\mu_0 + \Sigma^{-1}{\textstyle \sum_{n=1}^N}w_n
x_n\big)\label{eq:weightedgaussposterior}\\
\cov_w\left[f_n, f_m\right]  &=
\nicefrac{1}{2}\tr{\Psi^T\Psi} + \nu_m^T\Psi\nu_n,
\]
where $\Sigma = QQ^T$, $\nu_n \defined Q^{-1}(x_n-\mu_w)$ and $\Psi\defined Q^{-1}\Sigma_wQ^{-T}$.
Thus the greedy selection and weight update can be performed without Monte Carlo estimation. 
We set $\Sigma_0 = \Sigma = I$, $\mu_0 = 0$, $d=200$, and $N=1,000$.
We used a learning rate of $\gamma_t = t^{-1}$, $T=100$ 
weight update optimization iterations, and $M = 200$ greedy iterations,
although note that this is an upper bound on the size of the coreset as the same data point
may be selected multiple times. 
The results in \cref{fig:gaussian} demonstrate that the use of a fixed weighting function $\hpi$ (and thus, a fixed tangent plane on the coreset manifold) 
fundamentally limits the quality coresets via past algorithms. In contrast, the proposed greedy algorithm is ``manifold-aware'' and is able to continually
improve the approximation, resulting in orders-of-magnitude improvements in KL divergence to the true posterior.

\begin{figure}[t!]
\begin{center}
\begin{subfigure}{0.35\textwidth}
\centering\includegraphics[width=\textwidth]{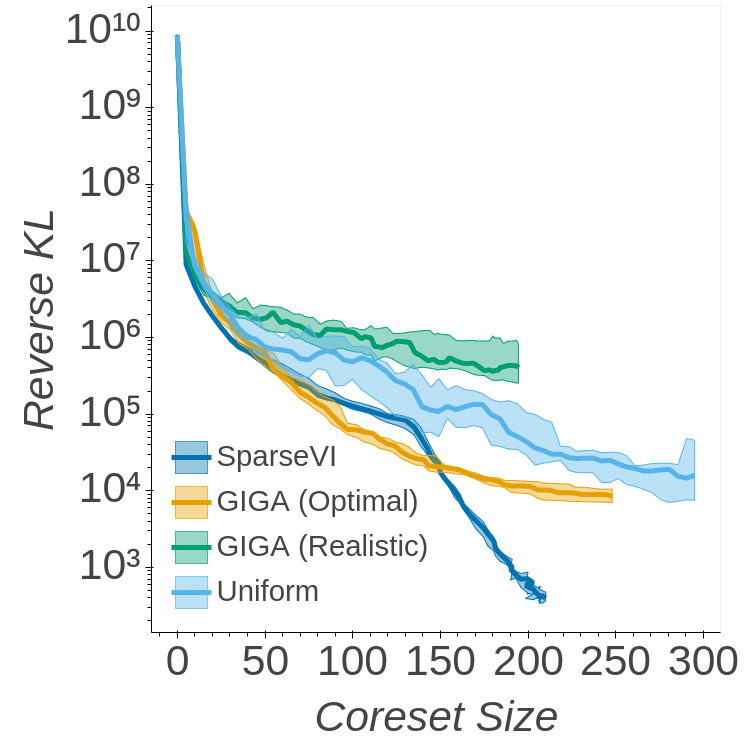}
 \caption{}\label{fig:linreg_kl}
\end{subfigure}
\begin{subfigure}{0.64\textwidth}
\centering\includegraphics[width=\textwidth]{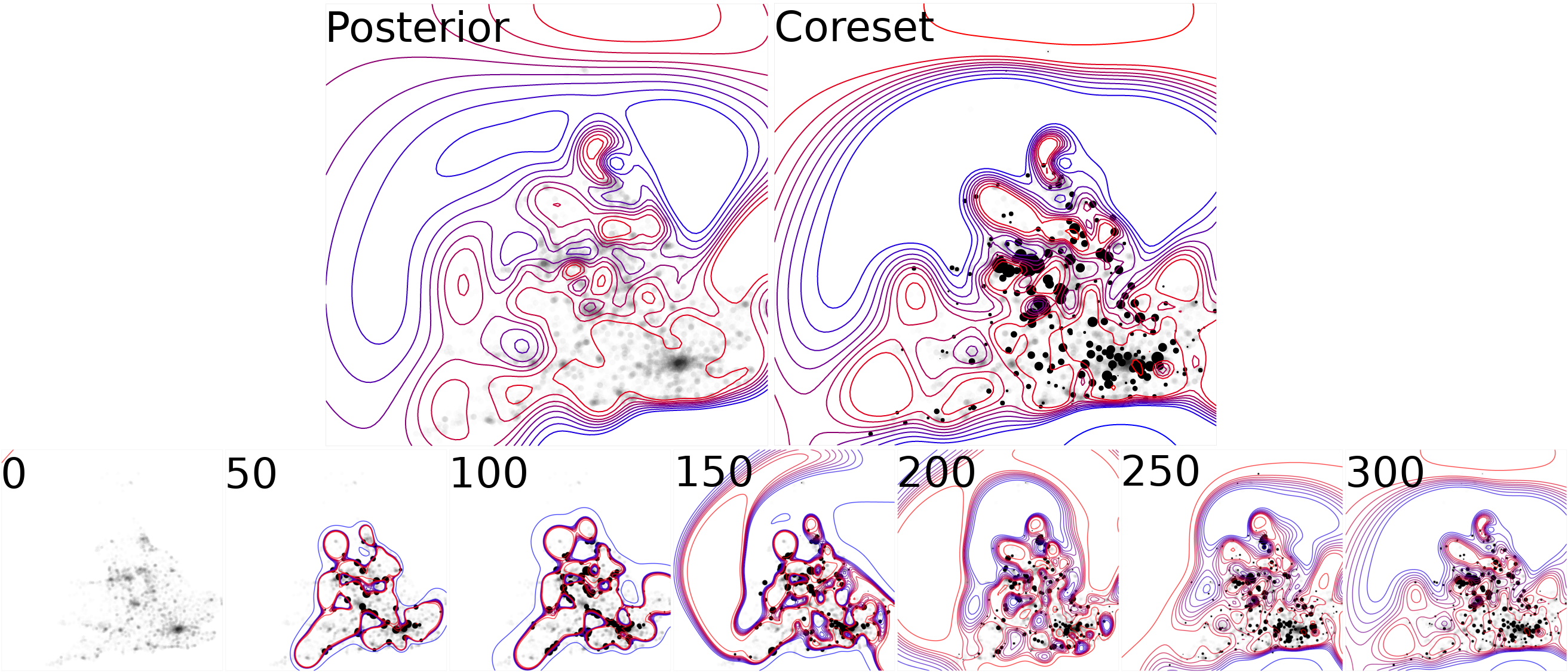}
 \caption{}\label{fig:linreg_coreset}
\end{subfigure}
\end{center}
\caption{(\ref{fig:linreg_kl}): Coreset construction on regression of housing prices using radial basis functions in the UK Land Registry data.
Solid lines show the median KL divergence over 10 trials, with $25^\text{th}$ and $75^\text{th}$ percentiles shown by shaded areas. 
(\ref{fig:linreg_coreset}): Posterior mean contours with coresets of size 0--300 via \texttt{SparseVI} compared with the exact posterior.
Posterior and final coreset highlighted in the top row. Coreset points are black with radius denoting weight.}\label{fig:linreg}
\vspace{-.2cm}
\end{figure}

\subsection{Bayesian radial basis function regression}\label{sec:linreg}
Next, we compared the coreset construction algorithms on Bayesian basis function regression for $N=10,000$ records of house sale
log-price $y_n\in\reals$ as a function of latitude / longitude coordinates $x_n\in\reals^2$
in the UK.\footnote{This dataset was constructed by merging housing prices from the UK land registry data \url{https://www.gov.uk/government/statistical-data-sets/price-paid-data-downloads}
with latitude \& longitude coordinates from the Geonames postal code data \url{http://download.geonames.org/export/zip/}.} The regression problem
involved inference for the coefficients $\alpha\in\reals^K$ in a linear combination of radial basis functions $b_k(x) = \exp(-\nicefrac{1}{2\sigma_k^2} (x - \mu_k)^2)$, $k=1,\dots,K$,
\[
y_n = b_n^T\alpha + \epsilon_n \quad \epsilon_n \distiid \distNorm(0, \sigma^2) \quad b_n = \left[\begin{array}{ccc}b_1(x_n) & \cdots & b_K(x_n)\end{array}\right]^T \quad \alpha \dist\distNorm(\mu_0, \sigma_0^2 I). 
\]
We generated 50 basis functions for each of 6 scales $\sigma_k \in \left\{0.2, 0.4, 0.8, 1.2, 1.6, 2.0\right\}$ by generating means $\mu_k$ uniformly from the data, and added one additional near-constant basis with scale 100 and mean corresponding to the mean latitude and longitude of the 
data. This resulted in $K=301$ total basis functions and thus a 301-dimensional regression problem. We set the prior and noise parameters $\mu_0, \sigma_0^2, \sigma^2$ equal to the empirical
mean, second moment, and variance of the price paid $(y_n)_{n=1}^N$ across the whole dataset, respectively.
As in \cref{sec:synthexpt}, the posterior and log-likelihood covariances are available in closed form, and all algorithmic steps can be performed without Monte Carlo.
In particular, $\pi_w = \distNorm(\mu_w, \Sigma_w)$, where (see \cref{sec:gaussiancovariance} for the derivation)
\[
\Sigma_w = (\Sigma_0^{-1} + \sigma^{-2}\sum_{n=1}^N &w_n b_nb_n^T)^{-1}\quad\text{and}\quad
\mu_w = \Sigma_w(\Sigma_0^{-1}\mu_0 + \sigma^{-2}\sum_{n=1}^N w_n y_n b_n)\\ \label{eq:weightedlinregposterior}
\cov_w\left[f_n, f_m\right] &=\sigma^{-4}\left(\nu_n\nu_m\beta_n^T\beta_m   +\nicefrac{1}{2}(\beta_n^T\beta_m)^2\right).
\]
where $\nu_n \defined y_n - \mu_w^Tb_n$, $\Sigma_w = LL^T$, and $\beta_n \defined L^Tb_n$.
We used a learning rate of $\gamma_t = t^{-1}$, $T = 100$ optimization steps, and $M = 300$ greedy iterations, although again note
that this is an upper bound on the coreset size. 
The results in \cref{fig:linreg} generally align with those from the previous synthetic experiment. The proposed sparse variational 
inference formulation builds coresets of comparable quality to Hilbert coreset construction (when given the exact posterior for $\hpi$)
 up to a size of about 150.
Beyond this point, past methods become limited by their fixed tangent plane approximation while the proposed method continues to improve. 
This experiment also highlights the sensitivity of past methods to the choice of $\hpi$: uniform subsampling
outperforms GIGA with a realistic choice of $\hpi$.

\subsection{Bayesian logistic and Poisson regression}
\begin{figure}[t!]
\begin{center}
\begin{subfigure}{0.48\textwidth}
\centering\includegraphics[width=\textwidth]{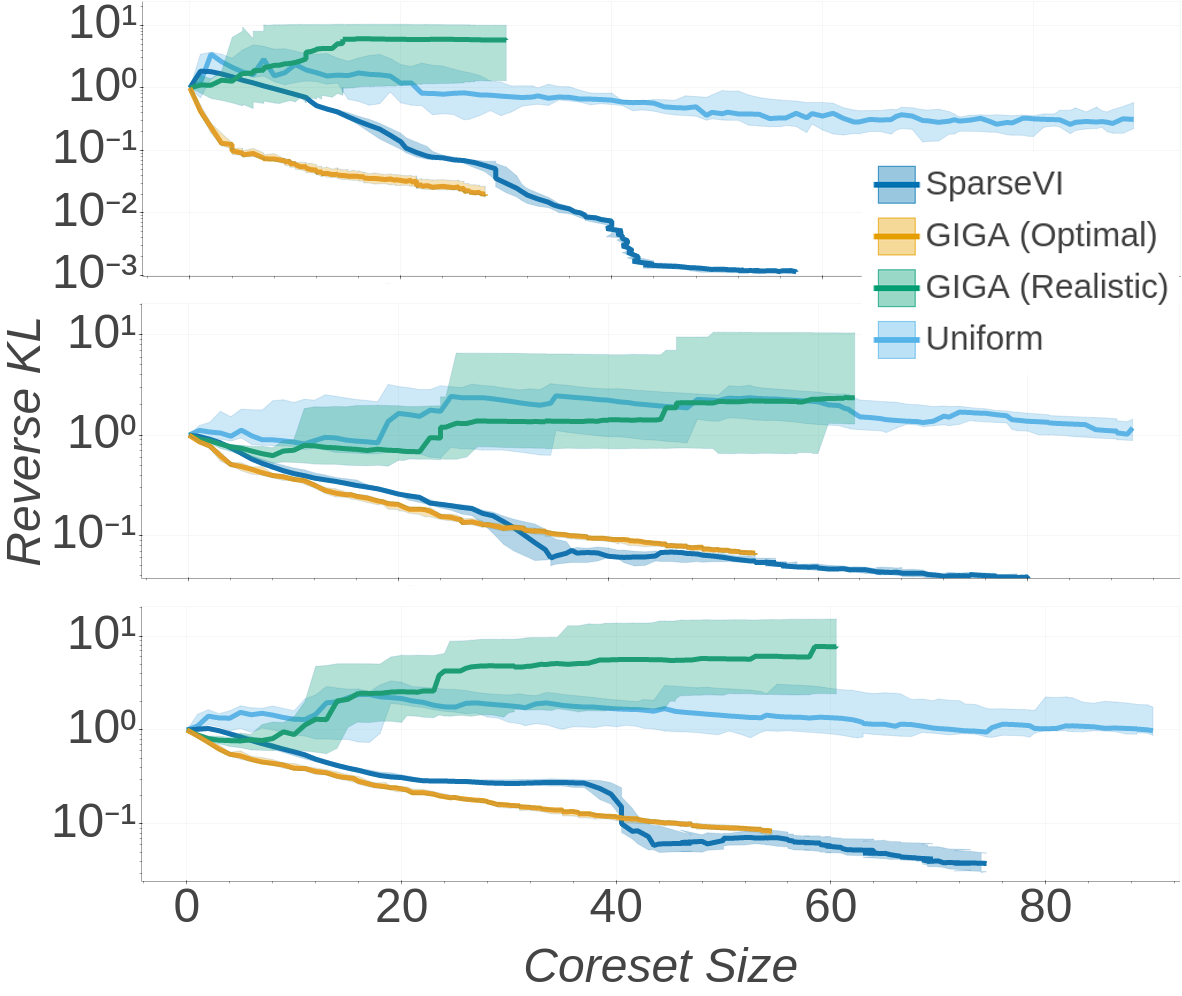}
 \caption{}\label{fig:lr_csz}
\end{subfigure}
\begin{subfigure}{0.48\textwidth}
\centering\includegraphics[width=\textwidth]{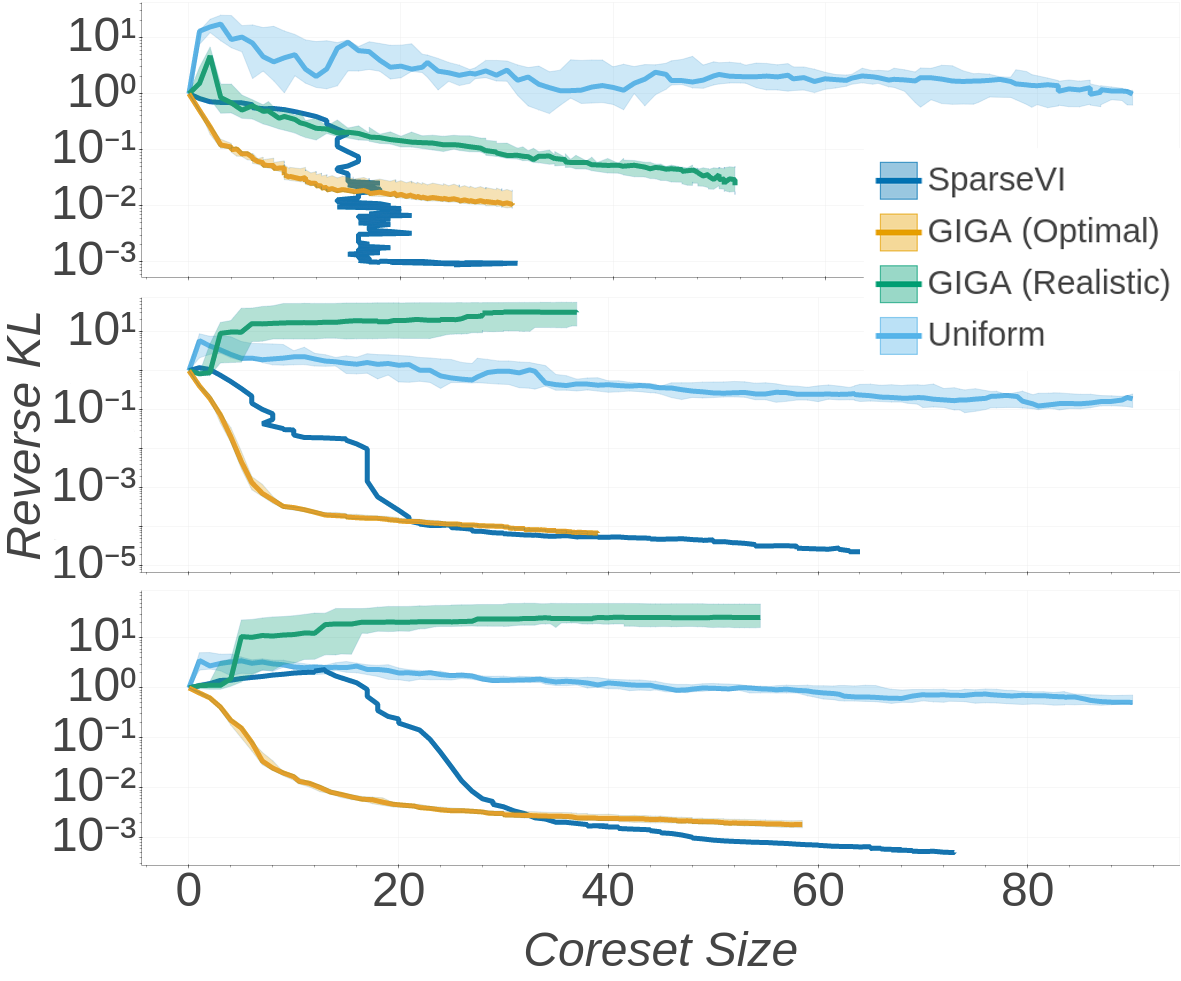}
 \caption{}\label{fig:pr_csz}
\end{subfigure}
\end{center}
\caption{The results of the logistic (\ref{fig:lr_csz}) and Poisson (\ref{fig:pr_csz}) regression experiments.
Plots show the median KL divergence (estimated using the Laplace approximation \citep{Tierney86} and normalized by the value for the prior) across 10 trials, with $25^\text{th}$ and $75^\text{th}$ percentiles shown by shaded areas.
From top to bottom, (\ref{fig:lr_csz}) shows the results for logistic regression on synthetic, chemical reactivities, and phishing websites data, 
while (\ref{fig:pr_csz}) shows the results for Poisson regression on synthetic, bike trips, and airport delays data. See \cref{sec:datasets} for details.}\label{fig:lrpoiss}
\end{figure}

Finally, we compared the methods on 
logistic and Poisson regression applied to six datasets (details may be found in \cref{sec:datasets}) with $N=500$ and dimension ranging from 2-15.
We used $M = 100$ greedy iterations, $S=100$ samples for Monte Carlo covariance estimation, and $T = 500$ optimization iterations with learning rate $\gamma_t = 0.5t^{-1}$. 
\cref{fig:lrpoiss} shows the result of this test, demonstrating that the proposed greedy sparse VI
method successfully recovers a coreset with divergence from the exact posterior as low or lower than GIGA with 
without having the benefit of a user-specified weighting function. Note that there is a computational price to pay for this level of automation;
\cref{fig:lrpoisstime}, \cref{sec:datasets} shows that \texttt{SparseVI} is significantly slower than 
Hilbert coreset construction via GIGA \citep{Campbell18}, primarily due to the expensive gradient descent weight update.
However, if we remove \texttt{GIGA (Optimal)} from consideration due to its unrealistic 
use of $\hpi\approx\pi_1$, \texttt{SparseVI} is the only practical coreset construction algorithm
that reduces the KL divergence to the posterior appreciably for reasonable coreset sizes. We leave improvements to computational cost for future work.

\section{Conclusion}
This paper introduced sparse variational inference for Bayesian coreset construction.
By exploiting the fact that coreset posteriors form an exponential family, a greedy algorithm
as well as a unifying Riemannian information-geometric view of present and past
coreset constructions were developed.
Future work includes extending sparse VI to improved optimization techniques
beyond greedy methods, and reducing computational cost.

\paragraph{Acknowledgments} T.~Campbell and B.~Beronov are supported by National Sciences and Engineering Research Council of Canada (NSERC) 
Discovery Grants. T.~Campbell is additionally supported by an NSERC Discovery Launch Supplement.

\small
\bibliographystyle{unsrt}
\bibliography{sources}

\begin{thebibliography}{10}

\bibitem{Robert04}
Christian Robert and George Casella.
\newblock {\em Monte Carlo Statistical Methods}.
\newblock Springer, 2nd edition, 2004.

\bibitem{Gelman13}
Andrew Gelman, John Carlin, Hal Stern, David Dunson, Aki Vehtari, and Donald
  Rubin.
\newblock {\em Bayesian data analysis}.
\newblock CRC Press, 3rd edition, 2013.

\bibitem{Jordan99}
Michael Jordan, Zoubin Ghahramani, Tommi Jaakkola, and Lawrence Saul.
\newblock An introduction to variational methods for graphical models.
\newblock {\em Machine Learning}, 37:183--233, 1999.

\bibitem{Wainwright08}
Martin Wainwright and Michael Jordan.
\newblock Graphical models, exponential families, and variational inference.
\newblock {\em Foundations and Trends in Machine Learning}, 1(1--2):1--305,
  2008.

\bibitem{Baydin18}
A{\i}l{\i}m~G{\"u}ne{\c{s}} Baydin, Barak Pearlmutter, Alexey Radul, and
  Jeffrey Siskind.
\newblock Automatic differentiation in machine learning: a survey.
\newblock {\em Journal of Machine Learning Research}, 18:1--43, 2018.

\bibitem{Kucukelbir17}
Alp Kucukelbir, Dustin Tran, Rajesh Ranganath, Andrew Gelman, and David Blei.
\newblock Automatic differentiation variational inference.
\newblock {\em Journal of Machine Learning Research}, 18:1--45, 2017.

\bibitem{Ranganath14}
Rajesh Ranganath, Sean Gerrish, and David Blei.
\newblock Black box variational inference.
\newblock In {\em International Conference on Artificial Intelligence and
  Statistics}, 2014.

\bibitem{Neal11}
Radford Neal.
\newblock {MCMC} using {H}amiltonian dynamics.
\newblock In Steve Brooks, Andrew Gelman, Galin Jones, and Xiao-Li Meng,
  editors, {\em Handbook of {M}arkov chain {M}onte {C}arlo}, chapter~5. CRC
  Press, 2011.

\bibitem{Hoffman14}
Matthew Hoffman and Andrew Gelman.
\newblock The {N}o-{U}-{T}urn {S}ampler: adaptively setting path lengths in
  {H}amiltonian {M}onte {C}arlo.
\newblock {\em Journal of Machine Learning Research}, 15:1351--1381, 2014.

\bibitem{Bardenet15}
R\'emi Bardenet, Arnaud Doucet, and Chris Holmes.
\newblock On {M}arkov chain {M}onte {C}arlo methods for tall data.
\newblock {\em Journal of Machine Learning Research}, 18:1--43, 2017.

\bibitem{Scott16}
Steven Scott, Alexander Blocker, Fernando Bonassi, Hugh Chipman, Edward George,
  and Robert McCulloch.
\newblock Bayes and big data: the consensus {M}onte {C}arlo algorithm.
\newblock {\em International Journal of Management Science and Engineering
  Management}, 11:78--88, 2016.

\bibitem{Betancourt15}
Michael Betancourt.
\newblock The fundamental incompatibility of {H}amiltonian {M}onte {C}arlo and
  data subsampling.
\newblock In {\em International Conference on Machine Learning}, 2015.

\bibitem{Alquier18}
Pierre Alquier and James Ridgway.
\newblock Concentration of tempered posteriors and of their variational
  approximations.
\newblock {\em The Annals of Statistics}, 2018 (to appear).

\bibitem{Wang18}
Yixin Wang and David Blei.
\newblock Frequentist consistency of variational {B}ayes.
\newblock {\em Journal of the American Statistical Association}, 0(0):1--15,
  2018.

\bibitem{Yang18}
Yun Yang, Debdeep Pati, and Anirban Bhattacharya.
\newblock $\alpha$-variational inference with statistical guarantees.
\newblock {\em The Annals of Statistics}, 2018 (to appear).

\bibitem{CheriefAbdellatif18}
Badr-Eddine Ch\'erief-Abdellatif and Pierre Alquier.
\newblock Consistency of variational {B}ayes inference for estimation and model
  selection in mixtures.
\newblock {\em Electronic Journal of Statistics}, 12:2995--3035, 2018.

\bibitem{Murphy12}
Kevin Murphy.
\newblock {\em Machine learning: a probabilistic perspective}.
\newblock The MIT Press, 2012.

\bibitem{Hoffman13}
Matthew Hoffman, David Blei, Chong Wang, and John Paisley.
\newblock Stochastic variational inference.
\newblock {\em The Journal of Machine Learning Research}, 14:1303--1347, 2013.

\bibitem{Rabinovich15}
Maxim Rabinovich, Elaine Angelino, and Michael Jordan.
\newblock Variational consensus {M}onte {C}arlo.
\newblock In {\em Advances in Neural Information Processing Systems}, 2015.

\bibitem{Broderick13}
Tamara Broderick, Nicholas Boyd, Andre Wibisono, Ashia Wilson, and Michael
  Jordan.
\newblock Streaming variational {B}ayes.
\newblock In {\em Advances in Neural Information Processing Systems}, 2013.

\bibitem{Campbell15}
Trevor Campbell, Julian Straub, John~W. {Fisher III}, and Jonathan How.
\newblock Streaming, distributed variational inference for {B}ayesian
  nonparametrics.
\newblock In {\em Advances in Neural Information Processing Systems}, 2015.

\bibitem{Welling11}
Max Welling and Yee~Whye Teh.
\newblock Bayesian learning via stochastic gradient {L}angevin dynamics.
\newblock In {\em International Conference on Machine Learning}, 2011.

\bibitem{Ahn12}
Sungjin Ahn, Anoop Korattikara, and Max Welling.
\newblock {B}ayesian posterior sampling via stochastic gradient {F}isher
  scoring.
\newblock In {\em International Conference on Machine Learning}, 2012.

\bibitem{Bardenet14}
R{\'e}mi Bardenet, Arnaud Doucet, and Chris~C Holmes.
\newblock Towards scaling up {M}arkov chain {M}onte {C}arlo: an adaptive
  subsampling approach.
\newblock In {\em International Conference on Machine Learning}, pages
  405--413, 2014.

\bibitem{Korattikara14}
Anoop Korattikara, Yutian Chen, and Max Welling.
\newblock Austerity in {MCMC} land: cutting the {M}etropolis-{H}astings budget.
\newblock In {\em International Conference on Machine Learning}, 2014.

\bibitem{Maclaurin14}
Dougal Maclaurin and Ryan Adams.
\newblock Firefly {M}onte {C}arlo: exact {MCMC} with subsets of data.
\newblock In {\em Conference on Uncertainty in Artificial Intelligence}, 2014.

\bibitem{Srivastava15}
Sanvesh Srivastava, Volkan Cevher, Quoc Dinh, and David Dunson.
\newblock {WASP}: scalable {B}ayes via barycenters of subset posteriors.
\newblock In {\em International Conference on Artificial Intelligence and
  Statistics}, 2015.

\bibitem{Entezari16}
Reihaneh Entezari, Radu Craiu, and Jeffrey Rosenthal.
\newblock Likelihood inflating sampling algorithm.
\newblock {\em arXiv:1605.02113}, 2016.

\bibitem{Angelino16}
Elaine Angelino, Matthew Johnson, and Ryan Adams.
\newblock Patterns of scalable {B}ayesian inference.
\newblock {\em Foundations and Trends in Machine Learning}, 9(1--2):1--129,
  2016.

\bibitem{Huggins16}
Jonathan Huggins, Trevor Campbell, and Tamara Broderick.
\newblock Coresets for {B}ayesian logistic regression.
\newblock In {\em Advances in Neural Information Processing Systems}, 2016.

\bibitem{Campbell17}
Trevor Campbell and Tamara Broderick.
\newblock Automated scalable {B}ayesian inference via {H}ilbert coresets.
\newblock {\em Journal of Machine Learning Research}, 20(15):1--38, 2019.

\bibitem{Campbell18}
Trevor Campbell and Tamara Broderick.
\newblock {B}ayesian coreset construction via greedy iterative geodesic ascent.
\newblock In {\em International Conference on Machine Learning}, 2018.

\bibitem{Clarkson10}
Kenneth Clarkson.
\newblock Coresets, sparse greedy approximation, and the {F}rank-{W}olfe
  algorithm.
\newblock {\em ACM Transactions on Algorithms}, 6(4), 2010.

\bibitem{LacosteJulien15}
Simon Lacoste-Julien and Martin Jaggi.
\newblock On the global linear convergence of {F}rank-{W}olfe optimization
  variants.
\newblock In {\em Advances in Neural Information Processing Systems}, 2015.

\bibitem{Locatello17}
Francesco Locatello, Michael Tschannen, Gunnar R\"atsch, and Martin Jaggi.
\newblock Greedy algorithms for cone constrained optimization with convergence
  guarantees.
\newblock In {\em Advances in Neural Information Processing Systems}, 2017.

\bibitem{Barron08}
Andrew Barron, Albert Cohen, Wolfgang Dahmen, and Ronald De{V}ore.
\newblock Approximation and learning by greedy algorithms.
\newblock {\em The Annals of Statistics}, 36(1):64--94, 2008.

\bibitem{Chen10}
Yutian Chen, Max Welling, and Alex Smola.
\newblock Super-samples from kernel herding.
\newblock In {\em Uncertainty in Artificial Intelligence}, 2010.

\bibitem{Schapire90}
Robert Schapire.
\newblock The strength of weak learnability.
\newblock {\em Machine Learning}, 5(2):197--227, 1990.

\bibitem{Huszar12}
Ferenc Huszar and David Duvenaud.
\newblock Optimally-weighted herding is {B}ayesian quadrature.
\newblock In {\em Uncertainty in Artificial Intelligence}, 2012.

\bibitem{Freund97}
Yoav Freund and Robert Schapire.
\newblock A decision-theoretic generalization of on-line learning and an
  application to boosting.
\newblock {\em Journal of Computer and System Sciences}, 55:119--139, 1997.

\bibitem{Candes05}
Emmanuel Cand\`es and Terence Tao.
\newblock Decoding by linear programming.
\newblock {\em IEEE Transactions on Information Theory}, 51(12):4203--4215,
  2005.

\bibitem{Candes07}
Emmanual Cand\`es and Terence Tao.
\newblock The {D}antzig selector: statistical estimation when $p$ is much
  larger than $n$.
\newblock {\em The Annals of Statistics}, 35(6):2313--2351, 2007.

\bibitem{Donoho06}
David Donoho.
\newblock Compressed sensing.
\newblock {\em IEEE Transactions on Information Theory}, 52(4):1289--1306,
  2006.

\bibitem{Boche15}
Holger Boche, Robert Calderbank, Gitta Kutyniok, and Jan Vyb\'iral.
\newblock A survey of compressed sensing.
\newblock In Holger Boche, Robert Calderbank, Gitta Kutyniok, and Jan
  Vyb\'iral, editors, {\em Compressed Sensing and its Applications: MATHEON
  Workshop 2013}. Birkh\"auser, 2015.

\bibitem{Mallat93}
St\'ephane Mallat and Zhifeng Zhang.
\newblock Matching pursuits with time-frequency dictionaries.
\newblock {\em IEEE Transactions on Signal Processing}, 41(12):3397--3415,
  1993.

\bibitem{Chen89}
Sheng Chen, Stephen Billings, and Wan Luo.
\newblock Orthogonal least squares methods and their application to non-linear
  system identification.
\newblock {\em International Journal of Control}, 50(5):1873--1896, 1989.

\bibitem{Chen99}
Scott Chen, David Donoho, and Michael Saunders.
\newblock Atomic decomposition by basis pursuit.
\newblock {\em SIAM Review}, 43(1):129--159, 1999.

\bibitem{Tropp04}
Joel Tropp.
\newblock Greed is good: algorithmic results for sparse approximation.
\newblock {\em IEEE Transactions on Information Theory}, 50(10):2231--2242,
  2004.

\bibitem{Tibshirani96}
Robert Tibshirani.
\newblock Regression shrinkage and selection via the lasso.
\newblock {\em Journal of the Royal Statistical Society Series B},
  58(1):267--288, 1996.

\bibitem{Geppert17}
Leo Geppert, Katja Ickstadt, Alexander Munteanu, Jesn Quedenfeld, and Christian
  Sohler.
\newblock Random projections for {B}ayesian regression.
\newblock {\em Statistics and Computing}, 27:79--101, 2017.

\bibitem{Ahfock17}
Daniel Ahfock, William Astle, and Sylvia Richardson.
\newblock Statistical properties of sketching algorithms.
\newblock {\em arXiv:1706.03665}, 2017.

\bibitem{Agarwal05}
Pankaj Agarwal, Sariel Har-Peled, and Kasturi Varadarajan.
\newblock {Geometric approximation via coresets}.
\newblock {\em Combinatorial and computational geometry}, 52:1--30, 2005.

\bibitem{Langberg10}
Michael Langberg and Leonard Schulman.
\newblock Universal $\epsilon$-approximators for integrals.
\newblock In {\em Proceedings of the $21^{\text{st}}$ Annual {ACM}--{SIAM}
  Symposium on Discrete Algorithms}, pages 598--607, 2010.

\bibitem{Feldman11}
Dan Feldman and Michael Langberg.
\newblock A unified framework for approximating and clustering data.
\newblock In {\em Proceedings of the $43^{\text{rd}}$ Annual {ACM} Symposium on
  Theory of Computing}, pages 569--578, 2011.

\bibitem{Feldman13}
Dan Feldman, Melanie Schmidt, and Christian Sohler.
\newblock Turning big data into tiny data: constant-size coresets for
  $k$-means, pca and projective clustering.
\newblock In {\em Proceedings of the $24^{\text{th}}$ Annual {ACM}--{SIAM}
  Symposium on Discrete Algorithms}, pages 1434--1453, 2013.

\bibitem{Bachem17}
Olivier Bachem, Mario Lucic, and Andreas Krause.
\newblock Practical coreset constructions for machine learning.
\newblock {\em arXiv:1703.06476}, 2017.

\bibitem{Braverman16}
Vladimir Braverman, Dan Feldman, and Harry Lang.
\newblock New frameworks for offline and streaming coreset constructions.
\newblock {\em arXiv:1612.00889}, 2016.

\bibitem{Frank56}
Marguerite Frank and Philip Wolfe.
\newblock An algorithm for quadratic programming.
\newblock {\em Naval Research Logistics Quarterly}, 3:95--110, 1956.

\bibitem{Guelat86}
Jacques Gu\'elat and Patrice Marcotte.
\newblock Some comments on {W}olfe's `away step'.
\newblock {\em Mathematical Programming}, 35:110--119, 1986.

\bibitem{Jaggi13}
Martin Jaggi.
\newblock Revisiting {F}rank-{W}olfe: projection-free sparse convex
  optimization.
\newblock In {\em International Conference on Machine Learning}, 2013.

\bibitem{Amari16}
Shun ichi Amari.
\newblock {\em Information Geometry and its Applications}.
\newblock Springer, 2016.

\bibitem{Tierney86}
Luke Tierney and Joseph Kadane.
\newblock Accurate approximations for posterior moments and marginal densities.
\newblock {\em Journal of the American Statistical Association},
  81(393):82--86, 1986.

\end{thebibliography}

\newpage
\appendix
\section{Proofs of results in the Riemannian information geometry section}\label{sec:riemannproofs}
\bprfof{\cref{prop:riemann}}
By \cref{eq:riemannmetric}, we have that $G(\hw) = \cov_{\hw}\left[f\right]$, and so
\[
\EE_{\hpi}\left[\left(\sum_{n=1}^N(1-w_n)g_n\right)^2\right] &=
\EE_{\hpi}\left[\left((1-w)^T\left(f - \EE_{\hpi}\left[f\right]\right)\right)^2\right]\\
&=
(1-w)^T\EE_{\hpi}\left[\left(f - \EE_{\hpi}\left[f\right]\right)\left(f - \EE_{\hpi}\left[f\right]\right)^T\right](1-w)\\
&= (1-w)^TG(\hw)(1-w)\\
&= ( (1-\hw)-(w-\hw))^TG(\hw)((1-\hw)-(w-\hw))\\
&= (\xi_{\hw\to 1} - \xi_{\hw \to w})^TG(\hw)(\xi_{\hw\to 1}-\xi_{\hw \to w}), \label{eq:geoalign}
\]
yielding the first result. Next, note that
\[
\left\|\xi_{w\to 1} - \xi_{w\to w+t_n1_n}\right\|_w &= t^2_n 1_n^T G(w)1_n - 2 t_n 1_n^TG(w)(1-w) + (1-w)^TG(w)(1-w), 
\]
and minimizing over $t_n$ yields
\[
t^\star_n = \frac{1_n^TG(w)(1-w)}{1_n^TG(w)1_n} \quad\text{or}\quad t^\star_n =
\max\left\{0, \frac{1_n^TG(w)(1-w)}{1_n^TG(w)1_n}\right\}
\]
if $t_n$  is unconstrained or positive-constrained, respectively.
Substituting back into the norm and using the definition of norms and inner
products via the Riemannian metric $G(w)$,
\[
\left\|\dots\right\|_w &=\left\{\begin{array}{ll}
   \|1-w\|_w^2\left(1 - \left(\left<\frac{1_n}{\|1_n\|_w}, \frac{1-w}{\|1-w\|_w}\right>_w\right)^2\right) & t_n \in\reals \\
   \|1-w\|_w^2\left(1 - \left(\max\left\{0, \left<\frac{1_n}{\|1_n\|_w},
   \frac{1-w}{\|1-w\|_w}\right>_w\right\}\right)^2\right) & t_n>0
 \end{array}\right.
\]
Finally, expressing the inner product explicitly,
\[
\left<\frac{1_n}{\|1_n\|_w}, \frac{1-w}{\|1-w\|_w}\right>_w &=
\frac{\EE_w\left[\left(f_n-\EE_wf_n\right)\left(f-\EE_wf\right)^T(1-w)
\right]}{\sqrt{\EE_w\left[\left(f_n-\EE_wf_n\right)^2\right]
\EE_w\left[\left(\left(f-\EE_wf\right)^T(1-w)\right)^2\right]}}\\
&= \corr_w\left[f_n, (1-w)^Tf\right],\label{eq:corrjust}
\]
yielding the second result.
\eprfof

\bnlem\label{lem:klest}
Define the path $\gamma(t)=(1-t)w+t1$. Then
\[
\kl{\pi_w}{\pi} &= 2(1-w)^T\EE\left[G(\gamma(T))\right](1-w) \quad T\dist\distBeta(1,2)\\
\kl{\pi}{\pi_w} &= 2(1-w)^T\EE\left[G(\gamma(S))\right](1-w)\quad S\dist\distBeta(2,1)\\
\kl{\pi_w}{\pi}+\kl{\pi}{\pi_w} &= (1-w)^T\EE\left[G(\gamma(U))\right](1-w)\quad U\dist\distUnif[0,1].
\]
\enlem
\bprf
Here we use prime notation for univariate differentiation.
For any twice differentiable function $h:[0,1] \to \reals$,
the Taylor remainder theorem states that
\[
h(1) &= h(0) + h'(0) + \int_0^1 h''(t)(1-t)\dee t.
\]
Let $\gamma : [0, 1] \to \reals_{\ge 0}^N$ be any twice-differentiable path
satisfying
$\gamma(0) = w$, $\gamma(1) = 1$, $\gamma'(0) = \gamma'(1) = 1-w$. Then if the log partition $\log Z(w)$ is also twice differentiable,
setting $h(t) = \log Z(\gamma(t))$ shows that
\[
\log Z(1) &= \log Z(w) + (1-w)^T\grad\log Z(w) + \nonumber\\
&\int_0^1 (1-t) \left(\gamma'(t)^T\grad^2\log Z(\gamma(t))\gamma'(t) + \gamma''(t)^T\grad\log Z(\gamma(t))\right)\dee t.
\]
Substituting into \cref{eq:sparsekl2} yields
\[
\kl{\pi_w}{\pi} &= \int_0^1 (1-t) \left(\gamma'(t)^T\grad^2\log Z(\gamma(t))\gamma'(t) + \gamma''(t)^T\grad\log Z(\gamma(t))\right)\dee t.
\]
The same logic follows with $\kl{\pi}{\pi_w}$, using a path $\zeta$ from $1$ to $w$ with $\zeta'(0) = \zeta'(1) = w-1$.
So selecting the path $\zeta(t) = \gamma(1-t)$ and using the transformation of variables $t \to 1-s$,
\[
\kl{\pi}{\pi_w} &= \int_0^1 t \left(\gamma'(t)^T\grad^2\log Z(\gamma(t))\gamma'(t) + \gamma''(t)^T\grad\log Z(\gamma(t))\right)\dee t.
\]
Adding the two expressions together makes the $t$ and $1-t$ terms cancel, and noting that the densities $\propto t$ and $\propto 1-t$ are beta densities yields the stated result.
\eprf

\bprfof{\cref{prop:hilbertcoreseterrorguarantees}}
By \cref{lem:klest} we have that
\[
\kl{\pi}{\pi_w} + \kl{\pi_w}{\pi} &\leq (1-w)^T\int_0^1 G(\gamma(t))\dee t(1-w).
\]
Multiplying and dividing by $J_{\hpi}(w) = (1-w)^T\grad^2\log Z(\hw)(1-w)$ from \cref{eq:geoalign}, 
defining $v\defined \frac{\grad^2\log Z(\hw)^{1/2}(1-w)}{\left\|\grad^2\log Z(\hw)^{1/2}(1-w)\right\|}$,
and defining $\tG(t) = G(\hw)^{-1/2}G(\gamma(t))G(\hw)^{-1/2}$
yields
\[
\hspace{-.3cm}\kl{\pi_w}{\pi} &= J_{\hpi}(w) \left(\int_0^1 (1-t) v^T \tG(t)v \dee t\right) \leq J_{\hpi}(w) \left(\int_0^1 (1-t) \lambda_{\max}\left(\tG(t)\right) \dee t\right) .
\]
Likewise,
\[
\kl{\pi}{\pi_w} &\leq J_{\hpi}(w) \left(\int_0^1 t v^T \tG(t)v \dee t\right) \leq J_{\hpi}(w) \left(\int_0^1 t \lambda_{\max}\left(\tG(t)\right) \dee t\right).
\]
Adding these equations yields the stated result.
\eprfof

\section{Weighted posterior and sufficient statistic covariance derivations}\label{sec:gaussiancovariance}
\subsection{Simple Gaussian inference}
The log likelihood for datapoint $x_n$ is (dropping normalization constants)
\[
f_n(\theta) &= - \frac{1}{2}\left(x_n-\theta\right)^T\Sigma^{-1}\left(x_n-\theta\right),
\]
so the $w$-weighted log-posterior is (again, up to normalization constants)
\[
\theta^T\left(\Sigma_0^{-1}\mu_0 + \Sigma^{-1}\sum_{n=1}^Nw_nx_n\right) - \frac{1}{2}\theta^T\left(\Sigma_0^{-1}+\sum_{n=1}^Nw_n\Sigma^{-1}\right)\theta.
\]
Completing the square yields \cref{eq:weightedgaussposterior}.
The first moment of the log-likelihood under the coreset posterior $\theta\dist\distNorm(\mu_w, \Sigma_w)$ is:
\[
\EE_{w}\left[f_n(\theta)\right] &= - \frac{1}{2}\EE_{w}\left[\left(x_n-\theta\right)^T\Sigma^{-1}\left(x_n-\theta\right)\right]\\
&= - \frac{1}{2}\tr{\Sigma^{-1}\Sigma_w} -\frac{1}{2}\left(\mu_w-x_n\right)^T\Sigma^{-1}\left(\mu_w-x_n\right)\\
&= - \frac{1}{2}\tr{\Psi} -\frac{1}{2}\|\nu_n\|^2,
\]
where $\Psi = Q^{-1}\Sigma_wQ^{-T}$,  $\nu_n = Q^{-1}(x_n-\mu_w)$, 
and $Q$ is the Cholesky decomposition of $\Sigma$, i.e., $\Sigma = QQ^T$. 
Defining $z \dist\distNorm(0, \Psi)$, its second moment is
\[
\EE_{w}\left[f_n(\theta)f_m(\theta)\right] &= \frac{1}{4}\EE_w\left[\left(x_n-\theta\right)^T\Sigma^{-1}\left(x_n-\theta\right)\left(x_m-\theta\right)^T\Sigma^{-1}\left(x_m-\theta\right)\right]\\
&= \frac{1}{4}\EE_{w}\left[(z-\nu_n)^T(z-\nu_n)(z-\nu_m)^T(z-\nu_m)\right]
\]
and by expanding and ignoring odd-order terms (which have 0 expectation),
\[
&=\frac{1}{4}\EE_{w}\left[
z^Tz z^Tz  + z^Tz \nu_m^T\nu_m
 +4 z^T\nu_nz^T\nu_m
+\nu_n^T\nu_nz^Tz + \nu_n^T\nu_n\nu_m^T\nu_m\right]\\
&= \frac{1}{4}\left(\left(\tr{\Psi}\right)^2+2\tr{\Psi^T\Psi}+ \|\nu_m\|^2\|\nu_n\|^2+\left(\|\nu_m\|^2+\|\nu_n\|^2\right)\tr{\Psi} + 4\nu_m^T\Psi\nu_n\right).
\]
So therefore,
\[
\cov_w\left[f_n,f_m\right] &= \EE_w\left[f_n(\theta)f_m(\theta)\right] - \EE_w\left[f_n(\theta)\right]\EE_w\left[f_m(\theta)\right]\\
&= \nu_m^T\Psi\nu_n + \frac{1}{2}\tr{\Psi^T\Psi}.
\]

\subsection{Bayesian radial basis regression}
The log likelihood for datapoint $n$ is (dropping normalization constants)
\[
f_n(\alpha) &= -\frac{1}{2\sigma^2}\left(y_n-\alpha^Tb_n\right)^2,
\]
so the $w$-weighted log-posterior is (again, up to normalization constants)
\[
\alpha^T\left(\sigma_0^{-2}\mu_0 + \sigma^{-2}\sum_{n=1}^Nw_ny_n b_n\right) - \frac{1}{2}\alpha^T\left(\sigma_0^{-2}I+\sigma^{-2}\sum_{n=1}^N w_n b_nb_n^T\right)\alpha.
\]
Completing the square yields \cref{eq:weightedlinregposterior}. The first moment of the log-likelihood under the coreset posterior $\alpha\dist\distNorm(\mu_w, \Sigma_w)$ is:
\[
\EE_w\left[f_n(\alpha)\right] &= -\frac{1}{2\sigma^2}\EE\left[\left(y_n-\mu_w^Tb_n\right)^2 + \left(\mu_w -\alpha\right)^Tb_nb_n^T\left(\mu_w-\alpha\right)\right]\\
 &= -\frac{1}{2\sigma^2}\left(\nu_n^2 + \tr b_nb_n^T\EE\left[\left(\mu_w-\alpha\right)\left(\mu_w -\alpha\right)^T\right]\right)\\
 &= -\frac{1}{2\sigma^2}\left(\nu_n^2 + \|\beta_n\|^2\right).
\]
where $\nu_n \defined (y_n - \mu_w^Tb_n)$,
$\Sigma_w = LL^T$, and $\beta_n = L^Tb_n$. Defining $Z = L^{-1}\left(\alpha - \mu_w\right) \dist \distNorm(0, I)$, the second moment is
\[
\EE_w\left[f_n(\alpha)f_m(\alpha)\right] &= \frac{1}{4\sigma^4}\EE_w\left[ \left(y_n-\alpha^Tb_n\right)^2\left(y_m-\alpha^Tb_m\right)^2\right]\\
&= \frac{1}{4\sigma^4}\EE_w\left[ \left(\nu_n-Z^T\beta_n\right)^2\left(\nu_m-Z^T\beta_m\right)^2\right].
\]
 Expanding and ignoring odd-order terms which have expectation 0, 
\[
&= \frac{1}{4\sigma^4}\left(\nu^2_n\nu^2_m + \nu_n^2\|\beta_m\|^2 +4\nu_n\nu_m\beta_n^T\beta_m +\nu_m^2\|\beta_n\|^2 + \sum_{i,j} \beta_{ni}^2\beta_{mj}^2  + 2\beta_{ni}\beta_{mi}\beta_{nj}\beta_{mj}\right)\\
&= \frac{1}{4\sigma^4}\left(\nu^2_n\nu^2_m + \nu_n^2\|\beta_m\|^2 +4\nu_n\nu_m\beta_n^T\beta_m +\nu_m^2\|\beta_n\|^2 + \|\beta_n\|^2\|\beta_m\|^2 +2(\beta_n^T\beta_m)^2 \right).
\]
Therefore, the covariance is
\[
\cov_w\left[f_n,f_m\right] &= \EE_w\left[f_n(\alpha)f_m(\alpha)\right] - \EE_w\left[f_n(\alpha)\right]\EE_w\left[f_m(\alpha)\right]\\
&=\frac{1}{\sigma^4}\left(\nu_n\nu_m\beta_n^T\beta_m   +\frac{1}{2}(\beta_n^T\beta_m)^2\right).
\]

\section{Details of the Logistic / Poisson regression experiment}\label{sec:datasets}
\begin{figure}[t!]
\begin{center}
\begin{subfigure}{0.48\textwidth}
\centering\includegraphics[width=\textwidth]{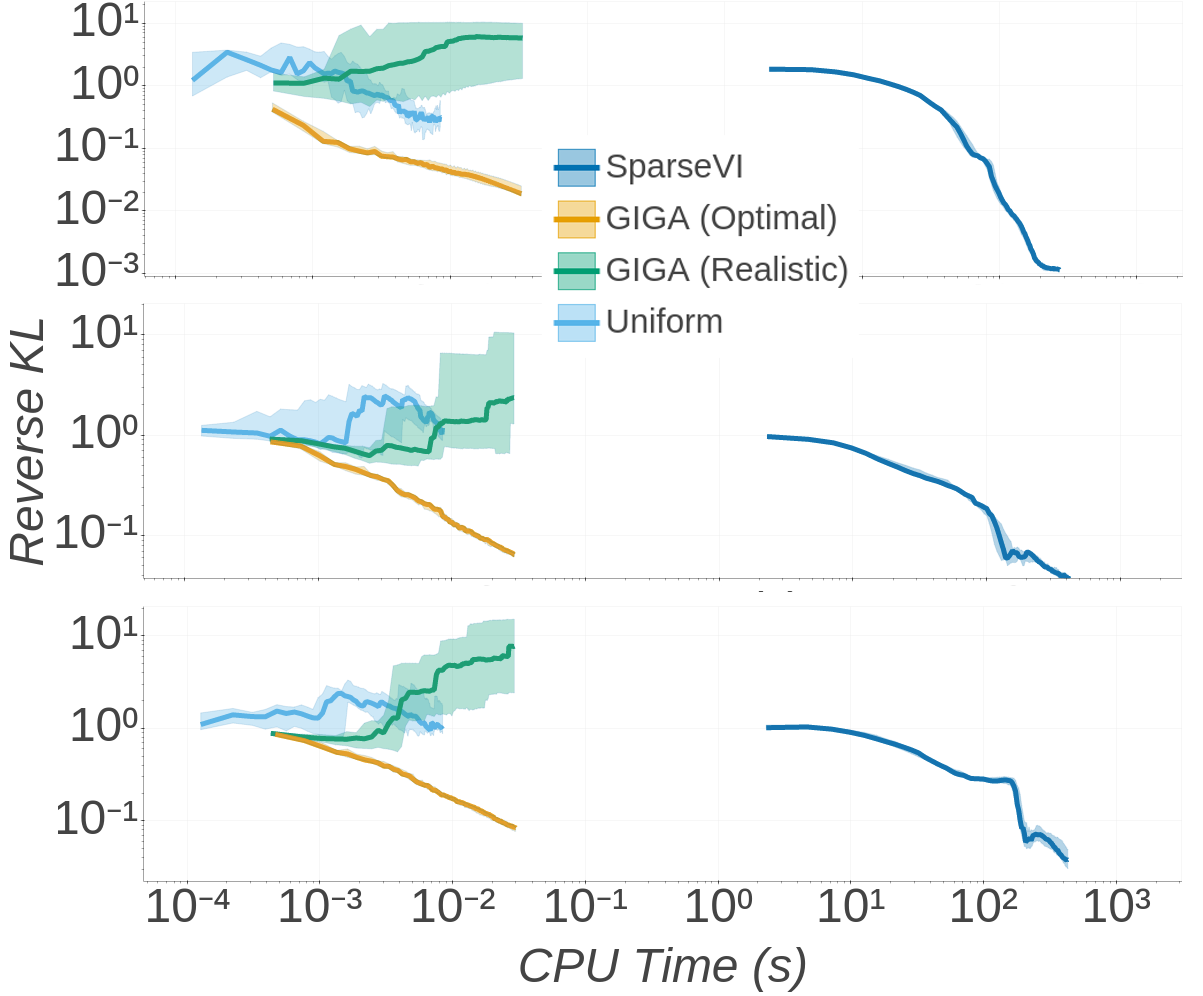}
 \caption{}\label{fig:lr_cput}
\end{subfigure}
\begin{subfigure}{0.48\textwidth}
\centering\includegraphics[width=\textwidth]{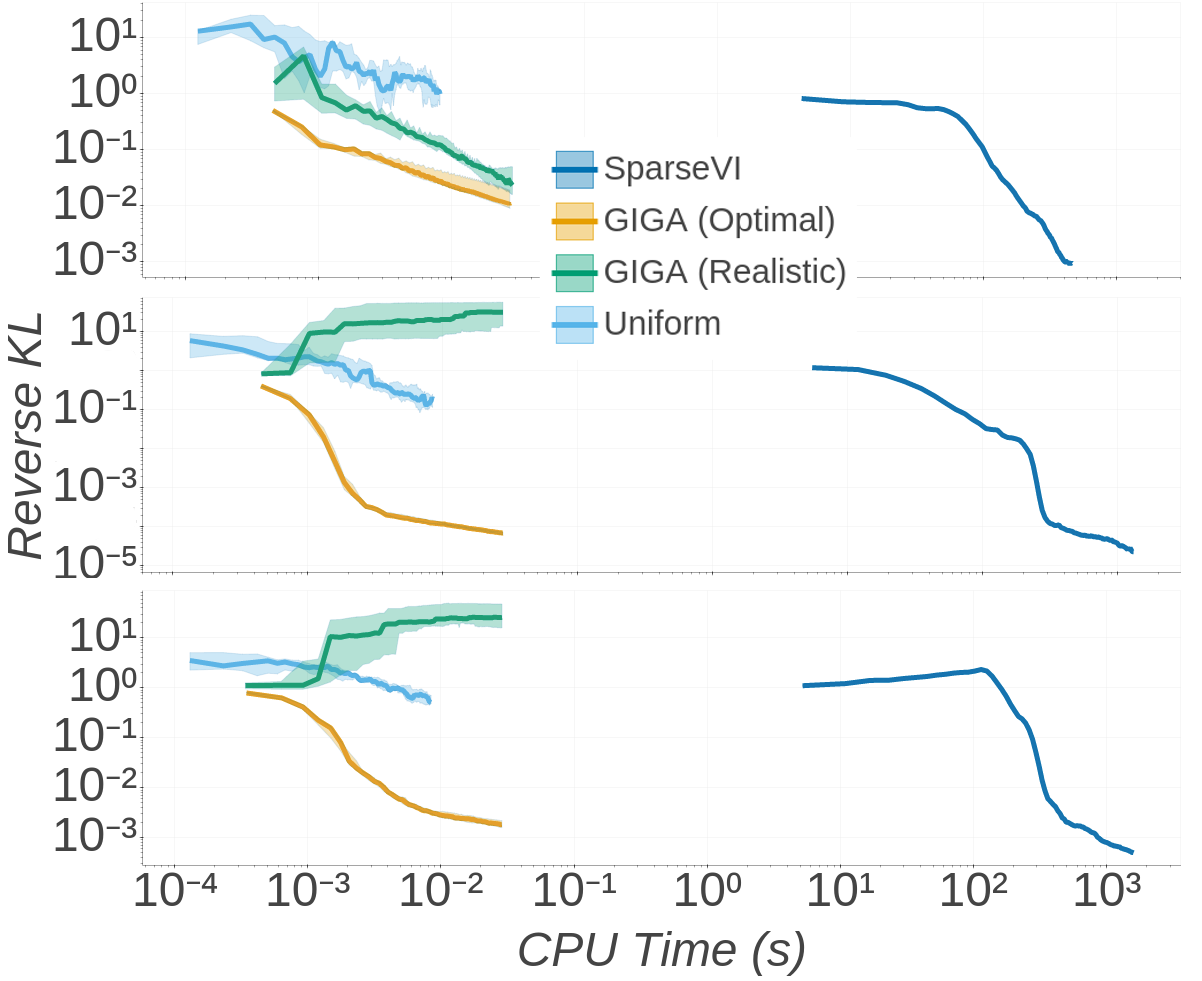}
 \caption{}\label{fig:pr_cput}
\end{subfigure}
\end{center}
\caption{Computation times for the logistic (\ref{fig:lr_cput}) and Poisson (\ref{fig:pr_cput}) regression experiments.
Plots show the median KL divergence (estimated using the Laplace approximation \citep{Tierney86} and normalized by the value for the prior) across 10 trials, with $25^\text{th}$ and $75^\text{th}$ percentiles shown by shaded areas.
From top to bottom, (\ref{fig:lr_cput}) shows the results for logistic regression on synthetic, chemical reactivities, and phishing websites data, 
while (\ref{fig:pr_cput}) shows the results for Poisson regression on synthetic, bike trips, and airport delays data.}\label{fig:lrpoisstime}
\end{figure}

In logistic regression, we are given a set of data points $\left(x_n, y_n\right)_{n=1}^N$ each consisting of a feature $x_n\in\reals^D$
and a label $y_n\in\{-1, 1\}$. The goal is to infer the posterior distribution of the latent parameter 
$\theta\in\reals^{D+1}$ in the following model:
\[
\hspace{-.3cm}y_n \given x_n, \theta &\distind
\distBern\left(\frac{1}{1+e^{-z_n^T\theta}}\right) \qquad z_n \defined
\left[\!\!\begin{array}{c}x_n\\ 1\end{array}\!\!\right].
\]
 We used three datasets (each subsampled to $N=500$ data points) in the logistic regression experiment:
a synthetic dataset with covariate $x_n \in \reals^2$ 
sampled \iid from $\distNorm(0, I)$, and label $y_n\in\{-1, 1\}$ generated from the logistic likelihood with parameter $\theta = \left[3, 3, 0\right]^T$;  
a phishing websites dataset reduced to $D=10$ features via principal component analysis;
and a chemical reactivity dataset with $D=10$ features.
The original phishing and chemical reactivities datasets are available online at \url{https://www.csie.ntu.edu.tw/~cjlin/libsvmtools/datasets/binary.html} and
 \url{http://komarix.org/ac/ds/}. Preprocessed versions for the experiments in this paper are available at \url{https://www.github.com/trevorcampbell/bayesian-coresets/}.

In Poisson regression, we are given a set of data points $\left(x_n,
y_n\right)_{n=1}^N$, each consisting of a feature $x_n\in\reals^D$
and a count $y_n\in\nats$.  The goal is to infer the posterior distribution of the latent parameter $\theta\in\reals^{D+1}$ in the following model:
\[
 \hspace{-.2cm}y_n \given x_n, \theta &\distind
 \distPoiss\left(\log\left(1+e^{z_n^T\theta}\right)\right) \qquad z_n \defined
 \left[\!\!\begin{array}{c}x_n\\ 1\end{array}\!\!\right].
\]
We used three additional datasets (each subsampled to $N=500$ data points) in the Poisson regression experiment:
a synthetic dataset 
with covariate $x_n \in \reals$ 
sampled \iid from $\distNorm(0, 1)$, and count $y_n\in\nats$ generated from the Poisson likelihood with $\theta = \left[1, 0\right]^T$;
a bikeshare dataset with $D=8$ features,
 relating the weather and seasonal information to the number of bike trips taken in an urban area;
and an airport delays dataset with $D=15$ features,
relating daily weather information to the number of flights leaving an airport
with a delay of more than 15 minutes.
The original bikeshare dataset is available online at \url{http://archive.ics.uci.edu/ml/datasets/Bike+Sharing+Dataset},
and the airport delays dataset was constructed using flight delay data from \url{http://stat-computing.org/dataexpo/2009/the-data.html}
and historical weather information from \url{https://www.wunderground.com/history/}. Preprocessed versions for the experiments in this paper are available at \url{https://www.github.com/trevorcampbell/bayesian-coresets/}.

\section{SparseVI optimization alternatives}\label{sec:alternativeupdate}
In the main text, we proposed one particular instantiation of sparse variational inference
based on a greedy iterative method and full gradient-descent-based weight update. There are
many possible variations on this theme; we highlight a few potential directions to explore
in future work below.

\subsection{Single weight update}
Rather than updating all the active weights, one might scale the current weights $w$ while adding the new component $1_{n^\star}$ via
\[
w^\star=\omega(\alpha^\star,\beta^\star) \quad \alpha^\star, \beta^\star =
\argmin_{\alpha,\beta \geq 0} \kl{\pi_{\omega(\alpha, \beta)}}{\pi} \quad
\text{s.t.} \quad \alpha,\beta \geq 0,
\label{eq:linesearch}
\]
where $\omega(\alpha, \beta) \defined \beta w+\alpha 1_{n^\star}$. To optimize, one would use Monte Carlo estimates of the gradients 
\[
\left[\begin{array}{c}\D{}{\beta}\\ \D{}{\alpha}\end{array}\right]
\kl{\pi_{\omega(\alpha,\beta)}}{\pi}&= \left[\begin{array}{cc}w &
1_{n^\star}\end{array}\right]^T \grad_{w}
\left.\kl{\pi_{w}}{\pi}\right|_{w=\omega(\alpha,\beta)}.
\label{eq:linesearchgrads}
\]

\subsection{Quadratic weight update}
The major computational cost in \texttt{SparseVI} is the weight updates in \cref{sec:weightupdate}:
for each gradient step, one must simulate a set of samples from $\pi_w$, compute all of the potentials,
and finally compute the Monte Carlo gradient estimate. Rather than optimizing the weights exactly, one might
minimize a quadratic expansion of the KL divergence at the point $w$,
\[
\!\!\!\!\kl{\pi_v}{\pi}\! \approx\! \kl{\pi_w}{\pi}\! +\!
(v\!-\!w)^T\grad_w\kl{\pi_w}{\pi} \!+\!
\frac{1}{2}(v\!-\!w)^T\grad_w^2\kl{\pi_w}{\pi}\! (v\!-\!w),\label{eq:quadratic}
\]
with Monte Carlo estimates of the gradient $D$ and Hessian $H$ 
based on the potential vector approximations $(\hg_s)_{s=1}^S$ already obtained
in the greedy selection step,
\[
\hspace{-.3cm}D &\defined -\frac{1}{S}\sum_{s=1}^S \hg_s\hg_s^T(1-w),\! & \!
LL^T = H &\defined \frac{1}{S}\sum_{s=1}^S \hg_s\hg_s^T(1-\hg_s^T(1-w)).
\]
Since \cref{eq:quadratic} is quadratic in $v$ or $\alpha,\beta$ (depending on
which type of weight update is used),
 the resulting weight update optimization is a nonnegative least squares
 problem,
\[
\hspace{-.2cm}v^\star &= \argmin_{v\in\reals^N,v\geq 0} \left\|L^T v\! -\!\left(L^T w - L^{-1} D\right)\right\|^2 
\,\,\text{s.t.}\,\,\left\{\begin{array}{ll}(1-1_{\mcI})^Tv=0 & \text{(fully corrective)} \\ v = \omega(\alpha,\beta) &  \text{(single-update)}\end{array}\right.\!\!\!\!\!.\label{eq:nnls}
\]
Upon solving the problem for $v^\star$, update the weights via $w\gets (1-\gamma_t)w + \gamma_t v^\star$ with a learning schedule $\gamma_t \geq 0$
to reduce the effect of Monte Carlo noise and aid in convergence.

\subsection{$\ell^1$-regularized coreset construction}\label{sec:l1}


Another option is to replace the cardinality constraint in \cref{eq:sparsekl}
with the standard $\ell^1$-norm regularization popularized by the LASSO method \citep{Tibshirani96} for sparse
linear regression,
\[
w^\star &= \argmin_{w\in\reals^N} \quad \kl{\pi_w}{\pi} + \lambda \tf^Tw \quad\text{s.t.}\quad w\geq 0,\label{eq:l1}
\]
with regularization weight $\lambda > 0$ and potential scales $\tf_n = \var_0{f_n}$. 
The potential scales $\tf$ account for the fact that the optimization is invariant to rescaling the potentials $f_n$ by positive constants;
the optimization \cref{eq:l1} is equivalent to optimizing $\kl{\pi_w}{\pi} + \lambda\|w\|_1$ with scale-invariant potentials $f_n/\sqrt{\var_0{f_n}}$.
We can solve this optimization for a particular value of $\lambda$
using proximal gradient descent,
\[
w_{t+1} \gets \prox{\gamma_t\lambda}{ w_t - \gamma_t \grad\kl{\pi_{w_t}}{\pi}},\,\,
\prox{\lambda}{x} &\defined 
\sgn(x)\max\left(|x|-\lambda\tf, 0\right),
\]
where $\gamma_t = O(1/t)$ is the learning rate when optimizing based on Monte Carlo estimates of $\grad\kl{\pi_w}{\pi}$. Although this approach 
generally provides less myopic solutions than greedy methods in the setting of sparse linear regression, 
there are two issues to address specific to sparse variational inference.
First, since estimating the gradient of the objective in \cref{eq:l1} involves sampling from $\pi_w$, the cost of iterations
increases as $w$ becomes dense. To avoid incurring undue cost, a binary search procedure on the regularization $\lambda$ may be used.
First, lower $\lambda_u$ and upper $\lambda_\ell$ bounds of $\lambda$ are initialized to 0 and $\max_n \left|\cov_0\left[f_n, f^T1\right]\right|$, respectively;
these bounds ensure that $\|w\|_0 = 0$ when $\lambda=\lambda_u$ and $\|w\|_0 = N$ when $\lambda=\lambda_\ell$.
Then in each binary search iteration optimization stage, keep track of $\|w\|_0$; if it ever becomes too large (e.g.~$2M$), return early
to prevent costly sampling steps.

\end{document}